%% file: optimalMP.tex
\documentclass[letterpaper, 10pt, conference]{./macros/ieeeconf}       
\IEEEoverridecommandlockouts                                             
\input{macros/packages.tex}

\newbool{conf}
\setbool{conf}{true}

\input{macros/commands.tex}
\input{macros/theorems.tex}

\begin{document}
\input{sections/title.tex}
\input{sections/abstract/main.tex}
\section{Introduction}
\input{sections/introduction/main.tex}

\section{Notation and Preliminaries}\label{section:preliminaries}
\input{sections/preliminaries/main.tex}
\section{Problem Statement}\label{section:problemstatements}
\input{sections/problemstatement/main.tex}

\section{HySST: An Optimal Motion Planning Algorithm for Hybrid Systems}\label{section:hysst}
\input{sections/algorithm/main.tex}

\section{Asymptotic Near-optimality Analysis}\label{section:mainresults}
\input{sections/mainresults/main.tex}

\section{HySST Software Tool for Optimal Motion Planning Problems for Hybrid Systems}\label{section:numerics}
\input{sections/numerics/main.tex}

\section{Conclusion}
\input{sections/conclusion/main.tex}
%\vspace{-0.3cm}
\bibliography{optimalMP}
\bibliographystyle{IEEEtran}

\end{document}

%% file: macros/packages.tex
\usepackage{./macros/hslTR}

\usepackage{graphics}
\usepackage{graphicx}
\usepackage{amssymb}
\usepackage{verbatim}     
\usepackage{amsxtra}
\usepackage{epsfig}
\usepackage{subfigure}
\usepackage{hyperref} 
\usepackage{amsmath}
\usepackage{latexsym}
\usepackage{ifthen}
\usepackage{psfrag}
\usepackage{subfigure}
\usepackage{color}
\usepackage{float}
\usepackage{tikz}
\usepackage{mathtools}
\usepackage{etoolbox}
\let\labelindent\relax
\usepackage{enumitem}
\usepackage{import}
\usepackage{./macros/rgsEnvironments}
\usepackage{./macros/rgsMacros}  

\usepackage{algorithm}
\usepackage{algpseudocode}

\usetikzlibrary{shapes,arrows}

%% file: macros/commands.tex
\newcommand{\pn}[1]{{#1}}

\newcommand{\Reals}[1][]{\mathbb{R}^{#1}}
\newcommand{\Preals}[1][]{\mathbb{R}^{#1}_{\geq 0}}
\newcommand{\Naturals}[1][]{\mathbb{N}^{#1}}
\newcommand{\StateofVertex}[1]{\overline{x}_{#1}}
\newcommand{\CostofVertex}[1]{\overline{c}_{#1}}
\newcommand{\Ballof}[2]{#2 + #1\mathbb{B}}
\newcommand{\Truncation}[3]{\widetilde{#1}_{(#2)\to(#3)}}

\newcommand{\Aux}{\overline{x}}
\newcommand{\Auf}{\overline{f}}
\newcommand{\Aug}{\overline{g}}
\newcommand{\Auc}{\overline{C}}
\newcommand{\Aud}{\overline{D}}
\newcommand{\Auh}{\overline{\mathcal{H}}}
\newcommand{\Auphi}{\overline{\phi}}
\newcommand{\Auxinit}{\overline{X}_{0}}
\newcommand{\Auxfinal}{\overline{X}_{f}}
\newcommand{\Auxunsafe}{\overline{X}_{u}}

\ifbool{conf}{
}{
 \addtolength{\oddsidemargin}{-.875in}
	\addtolength{\evensidemargin}{-.875in}
	\addtolength{\textwidth}{1.75in}
	\addtolength{\topmargin}{-.875in}
	\addtolength{\textheight}{1.75in} 
}

%% file: macros/theorems.tex
 \newcounter{theorem}
\newtheorem{problem}[theorem]{Problem}

\DeclareMathOperator{\interior}{int}
\DeclareMathOperator*{\argmin}{arg\,min}

\newlist{steps}{enumerate}{1}
\setlist[steps, 1]{leftmargin=45pt, label = \textbf{Step \arabic*}:}

%% file: sections/title.tex
\ifbool{conf}{
% IF CONFERENCE
\title{\LARGE \bf
	HySST: A Stable Sparse Rapidly-Exploring Random Trees Optimal Motion Planning Algorithm for Hybrid Dynamical Systems*
}
\author{Nan Wang and Ricardo G. Sanfelice% <-this % stops a space
	\thanks{*Research partially supported by NSF Grants no. ECS-1710621, CNS-2039054, and CNS-2111688, by AFOSR Grants no. FA9550-19-1-0053, FA9550-19-1-0169, and FA9550-20-1-0238, and by ARO Grant no. W911NF-20-1-0253. }% <-this % stops a space
	\thanks{Nan Wang and Ricardo G. Sanfelice are with the Department of Electrical and Computer Engineering, University of California, Santa Cruz, CA 95064, USA;
		{\tt\small nanwang@ucsc.edu, ricardo@ucsc.edu}}%
}
\maketitle
\maketitle}{
% IF REPORT
%%%
%%% STEP 2
%%%
%%% Include the following code after the "\begin{document}"
%%% statement. Make sure you set the following:
%%%
%%% \ititle{}  : title of the paper
%%% \iauthor{} : authors of the paper
%%% \iyear{}   : year
%%% \iredno{}  : report number (make sure you get a unique report
%%%             number)
%%%
%%% -- - BEGIN ISG TR - --
\ititle{HySST: A Samping-based Optimal Motion Planning Algorithm for Hybrid Systems.}
\iauthor{
  Nan Wang \\
  {\normalsize nanwang@ucsc.edu} \\
    Ricardo Sanfelice \\
  {\normalsize ricardo@ucsc.edu}}
\idate{\today{}} 
\iyear{2023}
\irefnr{02}
\makeititle}

%% file: sections/abstract/main.tex
\begin{abstract}
	This paper proposes a stable sparse rapidly-exploring random trees (SST) algorithm to solve the optimal motion planning problem for hybrid systems. At each iteration, the proposed algorithm, called HySST, selects a vertex with the lowest cost among all the vertices within the neighborhood of a randomly selected sample and then extends the search tree by flow or jump, which is also chosen randomly when both regimes are possible. In addition, HySST maintains a static set of witness \pn{points} such that all the vertices within the neighborhood of each witness are pruned except the vertex with the lowest cost. Through a definition of concatenation of functions defined on hybrid time domains, we show that HySST is asymptotically near optimal, namely, the probability of failing to find a motion plan such that its cost is close to the optimal cost approaches zero as the number of iterations of the algorithm increases to \pn{infinity}. This property is guaranteed under mild conditions on the data defining the motion plan, which include a relaxation of the usual positive clearance assumption imposed in the literature of classical systems. The proposed algorithm is applied to an actuated bouncing ball system and a collision-resilient tensegrity multicopter system so as to highlight its generality and computational features.
\end{abstract}

%% file: sections/introduction/main.tex
Motion planning consists of finding a state trajectory and associated inputs that \pn{connect} the initial and final state sets while satisfying the dynamics of the systems and given safety requirements. Motion planning for purely continuous-time systems and purely discrete-time systems \pn{has} been well studied in the literature; see e.g., \cite{lavalle2006planning}. 
In recent years, \pn{several (feasible)} motion planning algorithms have been developed, \pn{including} graph search algorithms \cite{wilfong1988motion}, artificial potential field methods \cite{khatib1985real} and  sampling-based algorithms. The sampling-based algorithms \pn{have} drawn much attention in recent years because of \pn{their} fast exploration speed \pn{for} high dimensional problems and theoretical guarantees; \pn{specially,} probabilistic completeness, which means that the probability of failing to find a motion plan converges to zero, as the number of samples approaches infinity. Two \pn{popular} sampling-based algorithms are the probabilistic roadmap (PRM) algorithm \cite{kavraki1996probabilistic} and \pn{the} rapidly-exploring random tree  (RRT) algorithm \cite{lavalle2001randomized}. The PRM method \pn{relies on} the existence of a steering function \pn{returning} the solution of a two-point boundary value problem (\pn{TP}BVP). \pn{Unfortunately, solutions to TPBVPs are difficult to generate} for most dynamical systems. \pn{On the other hand,} RRT algorithm does not require a steering function. Arguably, RRT is \pn{perhaps} the most successful algorithm \pn{to solve} feasible motion planning problems.
%RRT algorithm incrementally constructs a tree of state trajectories toward random samples in the state space.
%Compared with the artificial potential field method, RRT is probabilistically complete \cite{kleinbort2018probabilistic} which means that the probability of failing to find a motion plan converges to zero, as the number of samples approaches infinity. 

A feasible solution is not sufficient in most applications \pn{as} the quality of the solution returned by the motion planning algorithms is \pn{key} \cite{yang2019survey}. 
%The optimality of the heuristic graph search algorithm, such as A* algorithm, is only guaranteed when the employed heuristics is admissible \cite{hart1968formal}, which is difficult to verify in practice. The optimal motion planning algorithms using APF are also developed. In \cite{vadakkepat2001application}, the APF method is combined with evolutionary algorithms to derive optimal potential field called evolutionary artificial potential field. However, there is no theoretical guarantee over the optimality of the solution when the system dynamics is considered. Arguably, the sampling-based optimal motion planning algorithms are most promising because of the success of the sampling-based algorithms in solving the feasible motion planning problems. 
It \pn{has been shown in \cite{nechushtan2010sampling}} that the solution returned by  RRT converges to a sub-optimal solution. Therefore, variants of PRM and RRT, such as PRM* and RRT* \cite{karaman2011sampling}, \pn{have been developed} to solve  optimal motion planning problems \pn{with guaranteed} asymptotic optimality. \pn{However,} both PRM* and RRT* require a steering function, which prevents them from being widely applied.  On the other hand, the stable sparse RRT (SST) \pn{algorithm} \cite{li2016asymptotically} \pn{does not require a steering function and} is guaranteed to be asymptotically near optimal, which means that the probability of finding a solution that has a cost close to the minimal cost \pn{converges} to one as the number of iterations goes to infinity. 

The aforementioned motion planning algorithms have been widely applied \pn{to} purely continuous-time and purely discrete-time systems. However, \pn{much less efforts have been devoted to} the motion planning for hybrid systems. In our previous work \cite{wang2022rapidly}, a feasible motion planning problem is formulated \pn{for hybrid system given in terms of}  hybrid equations as in \cite{sanfelice2021hybrid}, which is a general framework that captures a broad class of hybrid systems. In \cite{wang2022rapidly}, \pn{the hybrid RRT} algorithm is designed to solve the feasible motion planning problem for \pn{such} systems with guaranteed probabilistic completeness. \pn{In this paper, we propose a motion planning algorithm for hybrid systems with the goal of assuring  (asymptotic) optimality of the solution.} We formulate the optimal motion planning problem for hybrid systems \pn{inspired by} the feasible motion planning problems in \cite{wang2022rapidly}. Then, we design an SST-type algorithm to solve the optimal motion planning problem for hybrid systems. Following \cite{li2016asymptotically}, the proposed algorithm, called HySST, incrementally constructs a search tree rooted in the initial state set toward the random samples. At first, HySST draws samples from the state space. Then, it selects \pn{a} vertex such that the state associated with this vertex is within a ball centered at the random sample and has minimal cost. Next, HySST propagates the state trajectory from the state associated with the selected vertex, and adds \pn{a} new vertex and edge from the propagated trajectory. In addition, HySST maintains a static set of state points, called witnesses, to represent the explored regions in the state space. HySST also employs a pruning process to guarantee that only a single vertex with lowest cost is kept within the neighborhood of each witness. \pn{We show that}, under mild assumptions, HySST is asymptotically near-optimal. To the authors' best knowledge, it is the first optimal RRT-type algorithm for  systems with hybrid dynamics. The proposed algorithm is illustrated in an actuated bouncing ball \pn{system} and a collision-resilient tensegrity multicopter system.

The remainder of the paper is structured as follows. Section \ref{section:preliminaries} presents notation and preliminaries. Section \ref{section:problemstatements} presents the problem statement and introduction of applications. Section \ref{section:hysst} presents the HySST algorithm. Section \ref{section:mainresults} presents the analysis of the asymptotic near optimality of HySST algorithm. Section \ref{section:numerics} presents the illustration of HySST in the examples. \ifbool{conf}{Proofs and more details are given \cite{nwang2023sst}.}{}

%% file: sections/preliminaries/main.tex
\subsection{Notation}
\input{sections/preliminaries/notations.tex}
\subsection{Preliminaries}
\input{sections/preliminaries/preliminaries.tex}

%% file: sections/preliminaries/notations.tex
The real numbers are denoted as $\Reals$ and its nonnegative subset is denoted as $\Preals$. The set of nonnegative integers is denoted as $\mathbb{N}$. The notation $\interior S$ denotes the interior of the set $S$. The notation $\overline{S}$ denotes the closure of the set $S$. The notation $\partial S$ denotes the boundary of the set $S$. The notation $\rge f$ denotes the range of the function $f$. The notation $\mathbb{B}$ denotes the closed unit ball of appropriate dimension in the Euclidean norm. The notation $\Ballof{r}{c}$ denotes the closed ball of appropriate dimension in the Euclidean norm centered at $c$ with radius $r$. Given sets $P\subset\mathbb{R}^n$ and $Q\subset\mathbb{R}^n$, the Minkowski sum of $P$ and $Q$, denoted as $P + Q$, is the set $\{p + q: p\in P, q\in Q\}$. \ifbool{conf}{}{The probability of the event $M$ is denoted as $Pr(M)$. } \ifbool{conf}{}{Given a set $S$, the notation $\mu(S)$ denotes its Lebesgue measure. The Lebesgue measure of the $n$-th dimensional unit ball, denoted $\zeta_{n}$, is such that
	\begin{equation}
	\label{equation:zetan}
	\zeta_{n} := \left\{\begin{aligned}
	&\frac{\pi^{k}}{k!} &\text{ if } n = 2k, k\in \mathbb{N}\\
	&\frac{2(k!)(4\pi)^{k}}{(2k + 1)!} &\text{ if } n = 2k + 1, k\in \mathbb{N}\\
	\end{aligned}\right.
	\end{equation}; see \cite{gipple2014volume}.}

%% file: sections/preliminaries/preliminaries.tex
A hybrid system $\mathcal{H}$ with inputs is modeled as  \cite{goebel2009hybrid}
\begin{equation}
\mathcal{H}: \left\{              
\begin{aligned}               
\dot{x} & = f(x, u)     &(x, u)\in C\\                
x^{+} & =  g(x, u)      &(x, u)\in D\\                
\end{aligned}   \right. 
\label{model:generalhybridsystem}
\end{equation}
where $x\in \mathbb{R}^n$ is the state, $u\in \mathbb{R}^m$ is the input, $C\subset \mathbb{R}^{n}\times\mathbb{R}^{m}$ represents the flow set, $f: \mathbb{R}^{n}\times\mathbb{R}^{m} \to \mathbb{R}^{n}$ represents the flow map, $D\subset \mathbb{R}^{n}\times\mathbb{R}^{m}$ represents the jump set, and $g:\mathbb{R}^{n}\times\mathbb{R}^{m} \to \mathbb{R}^{n}$ represents the jump map, respectively. The continuous evolution of $x$ is captured by the flow map $f$. The discrete evolution of $x$ is captured by the jump map $g$. The flow set $C$ collects the points where the state can evolve continuously. The jump set $D$ collects the points where jumps can occur.

Given a flow set $C$, the set $U_{C} := \{u\in \mathbb{R}^{m}: \exists x\in \mathbb{R}^{n}\text{ such that } (x, u)\in C\}$ includes all possible input values that can be applied during flows. Similarly, given a jump set $D$, the set $U_{D} := \{u\in \mathbb{R}^ {m}: \exists x\in \mathbb{R}^{n}\text{ such that } (x, u)\in D\}$ includes all possible input values that can be applied at jumps. These sets satisfy $C\subset \mathbb{R}^{n}\times U_{C}$ and $D\subset \mathbb{R}^{n}\times U_{D}$. Given a set $K\subset \mathbb{R}^{n}\times U_{\star}$, where $\star$ is either $C$ or $D$, we define
	$
	\Pi_{\star}(K) := \{x: \exists u\in U_{\star} \text{ s.t. } (x, u)\in K\}
	$
	as the projection of $K$ onto $\mathbb{R}^{n}$, and define \ifbool{conf}{$C' := \Pi_{C}(C)$ and $D' := \Pi_{D}(D)$.}{
		\begin{equation}
		\label{equation:Cprime}
		C' := \Pi_{C}(C)
		\end{equation} and 
		\begin{equation}
		\label{equation:Dprime}
		D' := \Pi_{D}(D).
		\end{equation}}

In addition to ordinary time $t\in \mathbb{R}_{\geq 0}$, we employ $j\in \mathbb{N}$ to denote the number of jumps of the evolution of $x$ and $u$ for $\mathcal{H}$ in (\ref{model:generalhybridsystem}), leading to hybrid time $(t, j)$ for the parameterization of its solutions and inputs. The domain of a solution to $\mathcal{H}$ is given by a hybrid time domain. A hybrid time domain is defined as a subset $E$ of $\mathbb{R}_{\geq 0}\times \mathbb{N}$ that, for each $(T, J)\in E$, $E\cap ([0, T]\times \{0, 1,..., J\})$ can be written as $\cup_{j = 0}^{J}([t_{j}, t_{j+1}],j)$ for some finite sequence of times $0=t_{0}\leq t_{1}\leq t_{2}\leq ... \leq t_{J+1} = T$. A hybrid arc $\phi: \dom \phi \to \mathbb{R}^{n}$ is a function on a hybrid time domain that, for each $j\in \mathbb{N}$, $t\mapsto \phi(t,j)$ is locally absolutely continuous on each interval $I^{j}:=\{t:(t, j)\in \dom \phi\}$ with nonempty interior. The definition of solution pair to a hybrid system is given as follows. For more details, see \cite{sanfelice2021hybrid}.
\begin{definition}[Solution pair to a hybrid system]
	\label{definition:solution}
	Given a pair of functions $\phi:\dom \phi\to \mathbb{R}^{n}$ and $u:\dom u\to \mathbb{R}^{m}$, $(\phi, u)$ is a solution pair to (\ref{model:generalhybridsystem}) if $\dom (\phi, u) := \dom \phi = \dom u$ is a hybrid time domain, $(\phi(0,0), u(0,0))\in \overline{C} \cup D$, and the following hold:
	\begin{enumerate}[label=\arabic*)]
		\item For all $j\in \mathbb{N}$ such that $I^{j}$ has nonempty interior, 
		\begin{enumerate}[label=\alph*)]
			\item the function $t\mapsto \phi(t, j)$ is locally absolutely continuous,
			\item $(\phi(t, j),u(t, j))\in C$ for all $t\in \interior I^j$,
			\item the function $t\mapsto u(t,j)$ is Lebesgue measurable and locally bounded,
			\item for almost all $t\in I^j$, \ifbool{conf}{$\dot{\phi}(t,j) = f(\phi(t,j), u(t,j))$.}{
				\begin{equation}
				\dot{\phi}(t,j) = f(\phi(t,j), u(t,j)).
				\end{equation}
			}
		\end{enumerate}
		\item For all $(t,j)\in \dom (\phi, u)$ such that $(t,j + 1)\in \dom (\phi, u)$, \ifbool{conf}{
			$$
			(\phi(t, j), u(t, j))\in D \quad
			\phi(t,j+ 1) = g(\phi(t,j), u(t, j)).
			$$}{
			\begin{equation}
			\begin{aligned}
			(\phi(t, j), u(t, j))&\in D\\
			\phi(t,j+ 1) &= g(\phi(t,j), u(t, j)).
			\end{aligned}
			\end{equation}
		}
	\end{enumerate}
\end{definition}

HySST requires concatenating solution pairs. The concatenation operation of solution pairs is defined next.
\begin{definition}
	\label{definition:concatenation}
	(Concatenation operation) Given two functions $\phi_{1}: \dom \phi_{1} \to \mathbb{R}^{n}$ and $\phi_{2}:\dom \phi_{2} \to \mathbb{R}^{n}$, where $\dom \phi_{1}$ and $\dom \phi_{2}$ are hybrid time domains, $\phi_{2}$ can be concatenated to $\phi_{1}$ if $ \phi_{1}$ is compact and $\phi: \dom \phi \to \mathbb{R}^n$ is the concatenation of $\phi_{2}$ to $\phi_{1}$, denoted $\phi = \phi_{1}|\phi_{2}$, namely,
	\begin{enumerate}[label=\arabic*)]
		\item $\dom \phi = \dom \phi_{1} \cup (\dom \phi_{2} + \{(T, J)\}) $, where $(T, J) = \max \dom \phi_{1}$ and the plus sign denotes Minkowski addition;
		\item $\phi(t, j) = \phi_{1}(t, j)$ for all $(t, j)\in \dom \phi_{1}\backslash \{(T, J)\}$ and $\phi(t, j) = \phi_{2}(t - T, j - J)$ for all $(t, j)\in \dom \phi_{2} + \{(T, J)\}$.
	\end{enumerate}
\end{definition}
\ifbool{conf}{}{The truncation operation is heavily used in proving the main results as is defined next.
\begin{definition}
	\label{definition: truncation}
	(Truncation and translation operation) Given a function $\phi: \dom \phi \to \mathbb{R}^{n}$, where $\dom \phi$ is hybrid time domain, and pairs of hybrid time $(T_{1}, J_{1})\in \dom \phi$ and $(T_{2}, J_{2})\in \dom \phi$ such that $T_{1}\leq T_{2}$ and $J_{1} \leq J_{2}$, the function, denoted $\Truncation{\phi}{T_{1}, J_{1}}{T_{2}, J_{2}}: \dom \Truncation{\phi}{T_{1}, J_{1}}{T_{2}, J_{2}} \to \mathbb{R}^{n}$, is the truncation of $\phi$ between $(T_{1}, J_{1})$ and $(T_{2}, J_{2})$ and translation by $(T_{1}, J_{1})$,  if
	\begin{enumerate}
		%	\item $\max_{t}\dom \widetilde\phi = T$, $\max_{j}\dom \widetilde\phi = J$;
		\item $\dom \Truncation{\phi}{T_{1}, J_{1}}{T_{2}, J_{2}} =  \dom \phi \cap ([T_{1}, T_{2}]\times \{J_{1}, J_{1} + 1, ..., J_{2}\}) - \{(T_{1}, J_{1})\}$, where the minus sign denotes Minkowski difference;
		\item $\Truncation{\phi}{T_{1}, J_{1}}{T_{2}, J_{2}}(t, j) = \phi(t + T_{1}, j + J_{1})$ for all $(t, j)\in \dom \widetilde\phi$.
	\end{enumerate}
\end{definition}}
\ifbool{conf}{}{In the main result of this paper, the following definition of closeness between hybrid arcs is used; see \cite{goebel2009hybrid}.
	\begin{definition}\label{definition:closeness}
		($(\tau, \epsilon)$-closeness of hybrid arcs) Given $\tau, \epsilon>0$, two hybrid arcs $\phi_{1}$ and $\phi_{2}$ are $(\tau, \epsilon)$-close if
		\begin{enumerate}
			\item for all $(t, j)\in \dom \phi_{1}$ with $t + j \leq \tau$, there exists $s$ such that $(s, j)\in \dom \phi_{2}$ and $|\phi_{1}(t, j) - \phi_{2}(s, j)| < \epsilon$;
			\item for all $(t, j)\in \dom \phi_{2}$ with $t + j \leq \tau$, there exists $s$ such that $(s, j)\in \dom \phi_{1}$ and $|\phi_{2}(t, j) - \phi_{1}(s, j)| < \epsilon$.
		\end{enumerate}
\end{definition}}

%% file: sections/problemstatement/main.tex
The feasible motion planning problem for hybrid systems is defined in \cite{wang2022rapidly} as follows.
\input{sections/problemstatement/feasiblemp.tex}
\input{sections/problemstatement/optimalmp.tex}

Problem \ref{problem:optimalmotionplanning} is illustrated in the following examples.
\input{sections/problemstatement/examplebb.tex}
\input{sections/problemstatement/examplecrtm.tex}

%% file: sections/problemstatement/feasiblemp.tex
\begin{problem} (Feasible motion planning)
	\label{problem:motionplanning}
	Given a hybrid system $\mathcal{H}$ with input $u\in \mathbb{R}^{m}$ and state $x\in \mathbb{R}^{n}$, the initial state set $X_{0}\subset\mathbb{R}^{n}$, the final state set $X_{f}\subset\mathbb{R}^{n}$, and the unsafe set $X_{u}\subset\mathbb{R}^{n}\times \mathbb{R}^{m}$, find a pair $(\phi, u): \dom (\phi, u)\to \mathbb{R}^{n}\times \mathbb{R}^{m}$, namely, \emph{a motion plan}, such that for some $(T, J)\in \dom (\phi, u)$, the following hold:
	\begin{enumerate}[label=\arabic*)]
		\item $\phi(0,0) \in X_{0}$, namely, the initial state of the solution belongs to the given initial state set $X_{0}$;
		\item $(\phi, u)$ is a solution pair to $\mathcal{H}$ as defined in Definition \ref{definition:solution};
		\item $(T,J)$ is such that $\phi(T,J)\in X_{f}$, namely, the solution belongs to the final state set at hybrid time $(T, J)$;
		\item $(\phi(t,j), u(t, j))\notin X_{u}$ for each $(t,j)\in \dom (\phi, u)$ such that $t + j \leq T+ J$, namely, the solution pair does not intersect with the unsafe set before its state trajectory reaches the final state set.
	\end{enumerate}
	Therefore, given sets $X_{0}$, $X_{f}$ and $X_{u}$, and a hybrid system $\mathcal{H}$ with data $(C, f, D, g)$, a (feasible) motion planning problem $\mathcal{P}$ is formulated as
	$
	\mathcal{P} = (X_{0}, X_{f}, X_{u}, (C, f, D, g)).
	$
\end{problem}

%% file: sections/problemstatement/optimalmp.tex
Let $\Sigma$ denote the set of all solution pairs to $\mathcal{H}$. Let $\Sigma_{\phi}$ denote the set of state trajectories of all the solution pairs in $\Sigma$. The optimal motion planning problem for hybrid systems \pn{consists of} finding a feasible motion plan with minimum cost \cite[Problem 3]{karaman2011sampling}.
\begin{problem}(Optimal motion planning)
	\label{problem:optimalmotionplanning}
	Given a motion planning problem $\mathcal{P} = (X_{0}, X_{f}, X_{u}, (C, f, D, g))$ and a cost functional $c: \Sigma_{\phi} \to \Preals$ where $\Sigma:= \{(\phi, u): (\phi, u) \text{ is a solution pair to } (C, f, D, g)\}$ and $\Sigma_{\phi} := \{\phi: \exists u: \dom \phi\to \Reals[m] \text{ such that } (\phi, u)\in \Sigma\}$, find a feasible motion plan $(\phi^{*}, u^{*})$ to Problem \ref{problem:motionplanning} such that $(\phi^{*}, u^{*}) = \argmin_{(\phi, u)\in \Sigma} c(\phi)$. Given sets $X_{0}$, $X_{f}$ and $X_{u}$, a hybrid system $\mathcal{H}$ with data $(C, f, D, g)$, and a cost functional $c$, an optimal motion planning problem $\mathcal{P}^{*}$ is formulated as
	$
	\mathcal{P}^{*} = (X_{0}, X_{f}, X_{u}, (C, f, D, g), c).
	$
\end{problem}

%% file: sections/problemstatement/examplebb.tex
\begin{example}\label{example:bouncingball}(Actuated bouncing ball system)
	Consider a ball bouncing on a fixed horizontal surface.
	% as is shown in Figure \ref{fig:bouncingball}. 
	The surface is located at the origin and, through control actions, is capable of affecting the velocity of the ball after the impact.  The dynamics of the ball while in the air is given by
	\ifbool{conf}{$\dot{x} = \left[ \begin{matrix}
			x_{2} \\
			-\gamma
		\end{matrix}\right] =: f(x, u),(x, u)\in C$}{\begin{equation}
		\label{model:bouncingballflow}
		\dot{x} = \left[ \begin{matrix}
		x_{2} \\
		-\gamma
		\end{matrix}\right] =: f(x, u)\qquad (x, u)\in C
		\end{equation}}
	where $x :=(x_{1}, x_{2})\in \mathbb{R}^2$. The height of the ball is denoted by $x_{1}$. The velocity of the ball is denoted by $x_{2}$. The gravity constant is denoted by $\gamma$. 
%	\begin{figure}
%		\centering
%		\incfig[0.27]{bouncingballfigure}
%		\caption{The actuated bouncing ball system\label{fig:bouncingball}.}
%	\end{figure}
\begin{figure}
	\centering
	\includegraphics[width=0.20\textwidth]{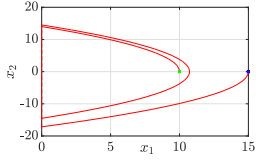}
	\caption{A motion plan to the sample motion planning problem for actuated  bouncing ball system.\label{fig:samplesolutionbb}}
\end{figure}
%	\begin{figure}[htbp] 
%		\centering
%		\subfigure[The actuated bouncing ball system\label{fig:bouncingball} ]{\incfig[0.27]{bouncingballfigure}}
%		\subfigure[A motion plan to the sample motion planning problem for actuated  bouncing ball system.\label{fig:samplesolutionbb}]{\includegraphics[width=0.20\textwidth]{figures/samplesolutionbb.eps}}
%		\caption{The actuated bouncing ball system in Example \ref{example:bouncingball}.}
%	\end{figure} 
	The flow is allowed when the ball is above the surface. Hence, the flow set is
	\ifbool{conf}{$C := \{(x, u)\in \mathbb{R}^{2}\times \mathbb{R}: x_{1}\geq 0\}.$}{\begin{equation}
		\label{model:bouncingballflowset}
		C := \{(x, u)\in \mathbb{R}^{2}\times \mathbb{R}: x_{1}\geq 0\}.
		\end{equation}}
	At every impact, the velocity of the ball changes from pointing down to pointing up while the height remains the same. The dynamics at jumps of the actuated bouncing ball system is given as
	\ifbool{conf}{$x^{+} = \left[ \begin{matrix}
		x_{1} \\
		-\lambda x_{2}+u
		\end{matrix}\right] =: g(x, u), (x, u)\in D$}{\begin{equation}
		x^{+} = \left[ \begin{matrix}
		x_{1} \\
		-\lambda x_{2}+u
		\end{matrix}\right] =: g(x, u)\qquad  (x, u)\in D
		\label{conservationofmomentum}
		\end{equation}}
	where $u\geq 0$ is the input and $\lambda \in (0,1)$ is the coefficient of restitution. 
	%\textcolor{red}{The surface has a constrained effect over the velocity after jumps. Let the lower bound over the input be $u_{\min}\geq 0$ and the upper bound over the input be $u_{\max}\geq u_{\min}$  such that $u_{\min}\leq u\leq u_{\max}$.   }
	The jump is allowed when the ball is on the surface with negative velocity. Hence, the jump set is
	\ifbool{conf}{$D:= \{(x, u)\in \mathbb{R}^{2}\times \mathbb{R}: x_{1} = 0, x_{2} \leq 0, u\geq 0\}.$}{\begin{equation}
		\label{model:bouncingballjumpset}
		D:= \{(x, u)\in \mathbb{R}^{2}\times \mathbb{R}: x_{1} = 0, x_{2} \leq 0, u\geq 0\}.
		\end{equation}}

	\ifbool{conf}{}{The hybrid model of the actuated bouncing ball system is given by  (\ref{model:generalhybridsystem}) where the flow map $f$ is given in (\ref{model:bouncingballflow}), the flow set $C$ is given in (\ref{model:bouncingballflowset}), the jump map $g$ is given in (\ref{conservationofmomentum}), and the jump set $D$ is given in (\ref{model:bouncingballjumpset}).}

	Given the initial state set $X_{0} = \{(15, 0)\}$, the final state set $X_{f} = \{(10, 0)\}$, and the unsafe set $X_{u} =\{(x, u)\in \mathbb{R}^{2}\times \mathbb{R}: x_1 \in [20, \infty), u\in[5, \infty)\}$, an instance of the optimal motion planning problem for the actuated bouncing ball system is to find a motion plan that has minimal hybrid time. To capture the hybrid time domain information, an auxiliary state $\tau\in \Reals_{\geq 0}$ representing the normal time and an auxiliary state $k \in \Naturals$ representing the jump numbers are imported. An auxiliary hybrid system $\Auh := (\Auc, \Auf, \Aud, \Aug)$ with state $\Aux := (x, \tau, k)\in \Reals[2]\times\Preals\times\Naturals$ is constructed as follows
	\ifbool{conf}{\begin{enumerate}
			\item $\Auc := \left\{(\Aux, u)\in \Reals[2]\times\Preals\times\Naturals\times\Reals: (x, u)\in C\right\}$;
			\item $\Auf(\Aux, u) := \left[\begin{matrix}
			f(x, u)\\
			1\\
			0
			\end{matrix}\right], (\Aux, u)\in \Auc$;
			\item $\Aud := \left\{(\Aux, u)\in \Reals[2]\times\Preals\times\Naturals\times\Reals: (x, u)\in D\right\}$;
			\item $\Aug(\Aux, u) := \left[\begin{matrix}
			g(x, u)\\
			\tau\\
			k + 1
			\end{matrix}\right], (\Aux, u)\in \Aud$.
	\end{enumerate}with the $X_{0}$, $X_{f}$, and $X_{u}$ extended to 
\begin{enumerate}
	\item $\Auxinit := X_{0}\times(0, 0)$;
	\item $\Auxfinal := X_{f}\times\Reals_{\geq 0}\times \Naturals$;
	\item $\Auxunsafe := X_{u}\times\Reals_{\geq 0}\times \Naturals$.
\end{enumerate}}{\begin{equation}
	\label{eq:auxsysflowset}
		\Auc := \left\{(\Aux, u)\in \Reals[2]\times\Preals\times\Naturals\times\Reals: (x, u)\in C\right\}
	\end{equation}
	\begin{equation}
		\label{eq:auxsysflowmap}
	\Auf(\Aux, u) := \left[\begin{matrix}
	f(x, u)\\
	1\\
	0
	\end{matrix}\right]\quad\forall (\Aux, u)\in \Auc
	\end{equation}
	\begin{equation}
		\label{eq:auxsysjumpset}
	\Aud := \left\{(\Aux, u)\in \Reals[2]\times\Preals\times\Naturals\times\Reals: (x, u)\in D\right\}
	\end{equation}
	\begin{equation}
			\label{eq:auxsysjumpmap}
	\Aug(\Aux, u) := \left[\begin{matrix}
	g(x, u)\\
	\tau\\
	k + 1
	\end{matrix}\right]\quad\forall (\Aux, u)\in \Aud
	\end{equation} with the sets $X_{0}$, $X_{f}$, and $X_{u}$ extended to 
	\begin{equation}\label{eq:auxsysinitial}
		\Auxinit := X_{0}\times(0, 0)
	\end{equation}
	\begin{equation}\label{eq:auxsysfinal}
	\Auxfinal := X_{f}\times\Reals_{\geq 0}\times \Naturals
	\end{equation}
	\begin{equation}\label{eq:auxsysunsafe}
	\Auxunsafe := X_{u}\times\Reals_{\geq 0}\times \Naturals
	\end{equation}}
	Then the cost functional $c$ can be defined as
	\begin{equation}\label{eq:auxsyscost}
		c(\Auphi) = c(\phi, \tau, k):= \tau(T,  J) + k(T,  J)
	\end{equation}
	where $\Auphi = (\phi, \tau, k)$ denotes a state trajectory of the solution pair to $\Auh$, and $(T, J) = \max \dom \Auphi$. Then the example optimal motion planning problem for the bouncing ball is defined as $\mathcal{P}^{*} = (\Auxinit, \Auxfinal, \Auxunsafe, (\Auc, \Auf, \Aud, \Aug), c)$
%	An instance of motion planning problem for the actuated bouncing ball system is given as follows.  The initial state set is . The final state set is $X_{f} = \{(10, 0)\}$. The unsafe set is $X_{u} =\{(x, u)\in \mathbb{R}^{2}\times \mathbb{R}: u\in[5, \infty)\}$.  The motion planning problem $\mathcal{P}$ is given as $\mathcal{P} = (X_{0}, X_{f}, X_{u}, (C, f, D, g))$. The state trajectory of this motion plan is shown in Figure \ref{fig:samplesolutionbb}. In the figure, the initial state set is denoted by a blue square. The final state set is denoted by a  green square. The red trajectory denotes the state trajectory of a sample motion plan.
\end{example}

%% file: sections/problemstatement/examplecrtm.tex
\begin{example}\label{example:multicopter} (Collision-resilient tensegrity multicopter system \cite{zha2021exploiting})
	Consider a collision-resilient tensegrity multicopter in horizontal plane that can operate after colliding with a concrete wall. The state of the multicopter is composed of the position vector $p := (p_{x}, p_{y})\in \Reals[2]$, where $p_{x}$ denotes the position along $x$-axle and $p_{y}$ denotes the position along $y$-axle, the velocity vector $v:= (v_{x}, v_{y})\in \Reals[2]$, where $v_{x}$ denotes the velocity along $x$-axle and $v_{y}$ denotes the velocity along $y$-axle, and the acceleration vector $a:= (a_{x}, a_{y})\in \Reals[2]$ where $a_{x}$ denotes the acceleration along $x$-axle and $a_{y}$ denotes the acceleration along $y$-axle. \pn{The} state \pn{of the system} is $x : = (p, v, a)\in \Reals[6]$ and \pn{its} input is $u :=(u_{x}, u_{y})\in \Reals[2]$.
	
	The environment is assumed \pn{to be} known. Define the region of the walls as  $\mathcal{W} \subset \Reals[2]$, represented by blue rectangles in Figure \ref{fig:drone}. \pn{Flow} is allowed when the multicopter is in the free space. Hence, the flow set is \ifbool{conf}{$C := \{((p, v, a), u)\in \Reals[6]\times \Reals[2]: p\notin \cal W\}.$}{\begin{equation}
		\label{model:crtmflowset}
		C := \{((p, v, a), u)\in \Reals[6]\times \Reals[2]: p\notin \cal W\}.
		\end{equation}}
%	\begin{figure}[htbp]
%		\centering
%		\includegraphics[width=0.8\columnwidth]{figures/crtm_map.eps}
%		\caption{The environment around the collision-resilient tensegrity multicopter. In the figure above, the blue rectangles denote the obstacles that may cause the collision with the multicopter. The cyan circle denotes the initial state and the magnet circle denotes the final state set.}\label{fig:crtmmap}
%	\end{figure}
The dynamics of the quadrotors when no collision occurs can be captured using  time-parameterized polynomial trajectories because of its differential flatness \ifbool{conf}{}{\cite{mellinger2012trajectory} }as
	\ifbool{conf}{$\dot{x} = \left[ \begin{matrix}
			v \\
			a\\
			u
		\end{matrix}\right] =: f(x, u), (x, u)\in C;$}{
	\begin{equation}
	\label{model:crtmflowmap}
	\dot{x} = \left[ \begin{matrix}
	v \\
	a\\
	u
	\end{matrix}\right] =: f(x, u), (x, u)\in C;
	\end{equation}}
	see \cite{liu2017search}. 

%	Let $v^{n}_{p, \cal S}$ denote the unit norm vector to the surface $\cal S$ of $\cal W$ at point $p$, assuming that $p\in \cal S$.
	 
%	At every impact, the velocity of the ball changes from pointing down to pointing up while the height remains the same. The dynamics at jumps of the actuated bouncing ball system is given as
Note that the post-collision position stays the same as the pre-collision position. Therefore, 
\ifbool{conf}{$p^{+} = p.$}{$$p^{+} = p.$$} Denote the velocity component of $v = (v_{x}, v_{y})$ that is normal to the wall as $v_{n}$ and the velocity component that is tangential to the wall as $v_{t}$. Then, the velocity component $v_{n}$ after the jump is modeled as\ifbool{conf}{$
	v_{n}^{+} = -ev_{n} =: g^{v}_{n}(v)
	$}{\begin{equation}\label{eq:vn}
		v_{n}^{+} = -ev_{n} =: g^{v}_{n}(v)
	\end{equation}}
 where $e\in (0, 1)$ is the coefficient of restitution.
The velocity component $v_{t}$ after the jump is modeled as
\ifbool{conf}{$v_{t}^{+} = v_{t} + \kappa(-e - 1)\arctan \frac{v_{t}}{v_{n}}v_{n} =: g^{v}_{t}(v)
	$}{\begin{equation}\label{eq:vt}
		v_{t}^{+} = v_{t} + \kappa(-e - 1)\arctan \frac{v_{t}}{v_{n}}v_{n} =: g^{v}_{t}(v)
	\end{equation}} where $\kappa\in\Reals$ is a constant; see \cite{zha2021exploiting}. Denote the projection of the updated vector $(v^{+}_{n}, v^{+}_{t})$ onto $x$ axle as $\Pi_{x}(v^{+}_{n}, v^{+}_{t})$ and the projection of the updated vector $(v^{+}_{n}, v^{+}_{t})$ onto $y$ axle as $\Pi_{y}(v^{+}_{n}, v^{+}_{t})$. Therefore,
$$
v^{+} = \left[\begin{matrix}
\Pi_{x}(g^{v}_{n}(v), g^{v}_{t}(v))\\
\Pi_{y}(g^{v}_{n}(v), g^{v}_{t}(v))
\end{matrix}\right] =: g^{v}(v).
$$

We assume that $a^{+} = 0$, which corresponds to a hovering status.
The discrete dynamics capturing the collision process is modeled as
\ifbool{conf}{$x^{+} = \left[ \begin{matrix}
		p \\
		g^{v}(v)\\
		0
	\end{matrix}\right] =: g(x, u),  (x, u)\in D.$}{
	\begin{equation}
	x^{+} = \left[ \begin{matrix}
	p \\
	g^{v}(v)\\
	0
	\end{matrix}\right] =: g(x, u)\qquad  (x, u)\in D
	\label{model:crtmjumpmap}
	\end{equation}}
	%\textcolor{red}{The surface has a constrained effect over the velocity after jumps. Let the lower bound over the input be $u_{\min}\geq 0$ and the upper bound over the input be $u_{\max}\geq u_{\min}$  such that $u_{\min}\leq u\leq u_{\max}$.   }
	The jump is allowed when the multicopter is on the wall surface with positive velocity towards the wall. Hence, the jump set is 
	\ifbool{conf}{$D:= \{((p, v, a), u)\in \mathbb{R}^{6}\times \mathbb{R}^{2}: p\in \partial \mathcal{W}, v_{n} < 0\}.$}{\begin{equation}
		\label{model:crtmjumpset}
		D:= \{((p, v, a), u)\in \mathbb{R}^{6}\times \mathbb{R}^{2}: p\in \partial \mathcal{W}, v_{n} < 0\}.
		\end{equation}}

	\ifbool{conf}{}{The hybrid model of the collision-resilient tensegrity multicopter system is given by  (\ref{model:generalhybridsystem}) where the flow map $f$ is given in (\ref{model:crtmflowmap}), the flow set $C$ is given in (\ref{model:crtmflowset}), the jump map $g$ is given in (\ref{model:crtmjumpmap}), and the jump set $D$ is given in (\ref{model:crtmjumpset}).}

	Given the initial state set as \pn{$X_{0} = \{(1,  2, 0, 0, 0, 0)\}$}, the final state set as $X_{f} = \{(5, 4)\}\times \Reals[4]$, and the unsafe set as $X_{u} =\{(x, u)\in \pn{\Reals[6]\times \Reals[2]}: p_x \in (\infty, 0]\cup[6, \infty), p_y \in (\infty, 0]\cup[5, \infty), (p_{x}, p_{y})\in \interior {\cal W}\}$, an instance of the optimal motion planning problem for the collision-resilient tensegrity multicopter system is to find the motion plan with minimal hybrid time \pn{as Example \ref{example:bouncingball}}.
\end{example}

%% file: sections/algorithm/main.tex
\input{sections/algorithm/overview.tex}
\input{sections/algorithm/algorithm.tex}

%% file: sections/algorithm/overview.tex
\subsection{Overview}
In this paper, we propose an SST algorithm for hybrid systems\pn{,} which we refer to as HySST algorithm. HySST algorithm searches for the optimal motion plan by incrementally constructing a search tree. The search tree is modeled by a directed tree. A directed tree is a pair $\mathcal{T} = (V, E)$, where $V$ is a set whose elements are called vertices and $E$ is a set of paired vertices whose elements are called edges. The edges in a directed tree are directed, which means the pairs of vertices that represent edges are ordered. The set of edges $E$ is defined as $E\subset \{(v_{1}, v_{2}): v_{1}\in V, v_{2}\in V, v_{1}\neq v_{2}\}$. The edge $e = (v_{1}, v_{2}) \in E$ represents an edge from $v_{1}$ to $v_{2}$. A path in $\mathcal{T}$ is a sequence of vertices $p = (v_{1}, v_{2},..., v_{k})$ such that $(v_{i}, v_{i+1})\in E$ for all $i \in \{1, 2,..., k - 1\}$. If there exists a path from the vertex $v_{0}$ to the vertex $v_{1}$ in the search tree, then $v_{0}$ is called a \emph{parent vertex} of $v_{1}$ and $v_{1}$ is called a \emph{child vertex} of $v_{0}$.

Each vertex in the search tree is associated with a state value of $\mathcal{H}$ and a cost value that estimates the cost from the root vertex up to the vertex. Each edge in the search tree is associated with a solution pair to $\mathcal{H}$ that connects the state values associated with their endpoint vertices. The state value and the cost value associated with vertex $v\in V$ is denoted as $\overline{x}_{v}$ and $\overline{c}_{v}$, respectively. The solution pair associated with edge $e\in E$ is denoted as $\overline{\psi}_{e}$. The solution pair that the path $p = (v_{1}, v_{2}, ..., v_{k})$ represents is the concatenation of all those solutions associated with the edges therein, namely,
\begin{equation}
\label{equation:concatenationpath}
\tilde{\psi}_{p} := \overline{\psi}_{(v_{1}, v_{2})}|\overline{\psi}_{(v_{2}, v_{3})}|\ ...\  |\overline{\psi}_{(v_{k-1}, v_{k})}	
\end{equation}
where $\tilde{\psi}_{p}$ denotes the solution pair associated with the path $p$. For the notion of concatenation, see Definition \ref{definition:concatenation}.

HySST requires a library of possible inputs. The input library $(\mathcal{U}_{C}, \mathcal{U}_{D})$ includes the input signals that can be applied during flows (collected in $\mathcal{U}_{C}$) and the input values that can be applied at jumps (collected in $\mathcal{U}_{D}$). \ifbool{conf}{}{To be more specific, HySST requires a library of inputs to simulate solution pairs. Note that inputs are constrained by the flow set $C$ and jump set $D$, respectively. Given the flow set $C$ and the jump set $D$ of the hybrid system $\mathcal{H}$, the input signal set during flows, denoted  $\mathcal{U}_{C}$, and the input values at jumps, denoted $\mathcal{U}_{D}$, are described as follows.
	\begin{enumerate}
		\color{red}
		\item The input signal applied during flows is a continuous-time signal. Therefore, the input signal during flows, denoted $\tilde{u}$, is specified by a function from an interval of time of the form $[0, t^{*}]$ to $U_{C}$ for some $t^{*}\in \mathbb{R}_{\geq 0}$, namely,
		$$
		\tilde{u}: [0, t^{*}]\to U_{C}.
		$$
		Definition \ref{definition:solution} also requires that $\tilde{u}$ is Lebesgue measurable and locally bounded.
		Then, the input signal set during flows, denoted $\mathcal{U}_{C}$, is a pre-determined set that collects possible $\tilde{u}$ that can be applied during flows. 
		
		Given $\tilde{u}\in\mathcal{U}_{C}$, the functional $\overline{t}: \mathcal{U}_{C} \to [0, \infty)$ returns the time duration of $\tilde{u}$. Namely, given any $\tilde{u}: [0, t^{*}]\to U_{C}$, $\overline{t}(\tilde{u}) = t^{*}$.
		\item The input applied at a single jump can be specified by an input value in $U_{D}$. The input set at jumps, denoted $\mathcal{U}_{D}$, is a pre-determined set that collects possible values of inputs that can be applied at jumps, namely, 
		$$
		\mathcal{U}_{D}\subset U_{D}.
		$$
	\end{enumerate}
	The following introduces a method to construct $\mathcal{U}_{C}$ and $\mathcal{U}_{D}$, given sets $C$ and $D$ respectively. The  procedure to construct $\mathcal{U}_{C}$ is given as follows.
	\begin{steps}
		\item Set $t^{*}$ to a positive constant. Choose a finite number of points from $U_C$ and denote it $U^s_C$.
		\item For each point $u^{s}\in U_{C}^{s}$, construct an input signal $[0, t^{*}]\to \{u^{s}\}$ and add it to $\mathcal{U}_{C}$. 
	\end{steps}
	Set $\mathcal{U}_{D}$ can be constructed  similarly by choosing a finite number of points from $U_{D}$.}

\ifbool{conf}{}{A key difference between HyRRT in \cite{wang2022rapidly} and HySST is that }HySST employs a pruning process to decrease the number of vertices in  the search tree. \ifbool{conf}{}{It is not required to keep all the vertices in the search tree because some of the vertices may be close to a vertex with much lower cost. This observation allows for a pruning operation to ignore some vertices generated during the search. }This pruning operation is implemented by maintaining a \emph{witness state set}, denoted as $S$, such that all the vertices within the vicinity of the witnesses are ignored except the one with lowest cost. For every witness $s$ kept in $S$, a single vertex in the tree represents that witness. Such a vertex is stored in $s.rep$ for each witness $s\in S$. 

HySST selects the vertex associated with the lowest cost within the vicinity of a randomly selected state. This vicinity of the randomly selected state, which we refer to as \emph{random state neighborhood}, is defined by a ball of radius $\delta_{BN}\in \mathbb{R}_{>0}$. The pruning process removes from the search tree all the vertices within the vicinity of  the closest witness to those removed vertices, which we refer to as \emph{closest witness neighborhood} and is defined by a ball of radius $\delta_{s}\in \mathbb{R}_{>0}$, The pruning process does not remove the vertices in the closest witness neighborhood with lowest cost. Note that a vertex, say, $v_{a}$, may be associated to a higher cost than the costs associated with other vertices within the closest witness neighborhood of $v_{a}$, but has a child vertex, say,  $v_{b}$, associated to the lowest cost compared with the costs of other vertices in the closest witness neighborhood of $v_{b}$. In this case, $v_{a}$ should not be removed from the search tree because if it is removed, $v_{b}$ is also removed due to being cascaded. However, even $v_{a}$ is not removed, it will not be selected, and, therefore, will be kept in a separate set called inactive vertex set, denoted $V_{inactive}$. On the other hand, the vertices that are not pruned are stored in a set called the active vertex set, denoted $V_{active}$. 
%\begin{figure}[htbp]
%	\centering
%	\incfig[0.8]{witnesssearchtree}
%	\caption{The search tree witnessed by a witness set. In this figure, the black dots and lines denote the active vertices and the edges in the search tree. The blue dots denote the inactive vertices. The red dots denote the witness states. The green dot denotes the randomly selected state $x_{rand}$. Any single witness has an active vertex as its representative.\label{fig:hysstoverview}}
%\end{figure}
%The pruning operation maintains vertices with lowest cost, denoted by the cost values associated with those vertices. This leads to a pruning method that evaluates the costs of the vertices in a local vicinity and prunes the vertices with higher costs in this vicinity as long as they do not have children with good path costs in their neighborhood. The pruned vertices will not be considered for propagation and are stored in the vertices set $V_{inactive}$. 
%The remaining vertices will still be considered in the following iterations for propagation and are stored in the vertices set $V_{active}$. 
%In order to define local neighborhoods, similar to SST, HySST uses an auxiliary set of states, called \emph{witnesses} and 

Next, we introduce the main steps executed by HySST. Given the optimal motion planning problem $\mathcal{P}^{*} = (X_{0}, X_{f}, X_{u}, (C, f, D, g), c)$ and the input library $(\mathcal{U}_{C}, \mathcal{U}_{D})$, HySST performs the following steps:
\begin{steps}
	\item Sample a finite number of points from $X_{0}$ and initialize a search tree $\mathcal{T} = (V, E)$ by adding vertices associated with each sampling point and setting $E$ as $E\leftarrow\emptyset$. 
	\item Initialize the witness state set $S\subset \Reals[n]$ by $S \leftarrow \emptyset$. For each $v\in V$ such that for all $ v' \in V\backslash v$, $|\overline{x}_{v} - \overline{x}_{v'}| > \delta_{s}$, add the witness state $s = \overline{x}_{v}$ to $S$ and set the representative of $s$ as $s.rep \leftarrow v$. Initialize the active vertices set $V_{active}$ by $V_{active}\leftarrow \{s.rep\in V: s\in S\}$. Initialize the inactive vertices set $V_{inactive}$ by $V_{inactive} \leftarrow\emptyset$. 
	\item Randomly select one regime among flow regime and jump regime for the evolution of $\mathcal{H}$.
	\item Randomly select a point $x_{rand}$ from $C'$ ($D'$) if the flow (jump, respectively) regime is selected in \textbf{Step 3}.
	\item Find all the vertices in $V_{active}$ associated with the state values that are within $\delta_{BN}$ to $x_{rand}$ and collect them in the set $V_{BN}$. Then, find the vertex in $V_{BN}$ that has minimal cost, denoted $v_{cur}$. If no vertex is collected in $V_{BN}$, then find the vertex in the search tree that has minimal distance to $x_{rand}$ and assign it to $v_{cur}$.
	%Denote the state value associated with $v_{cur}$ as $\overline{x}_{v_{cur}}$, as is shown in Figure \ref{fig:searchgraph_statespace}.	
	\item Randomly select an input signal (value) from $\mathcal{U}_{C}$ ($\mathcal{U}_{D}$) if $\overline{x}_{v_{cur}}\in C'\backslash D'$ ($\overline{x}_{v_{cur}}\in D'\backslash C'$). Then, compute a solution pair denoted $\psi_{new} = (\phi_{new}, u_{new})$ starting from $\overline{x}_{v_{cur}}$ with the selected input applied via flow (jump, respectively). If $\overline{x}_{v_{cur}}\in D'\cap C'$, a random process is employed to decide to proceed the computation with flow or jump.
	Denote the final state of $\phi_{new}$ as $x_{new}$. Compute the cost at $x_{new}$, denoted $c_{new}$, by $c_{new}\leftarrow \overline{c}_{v_{cur}} + c(\phi_{new})$. If $\psi_{new}$ does not intersect with $X_{u}$, then go to next step. Otherwise, go to \textbf{Step 3}.
	\item  Find one of the witnesses in $S$ that has minimal distance to $x_{new}$, denoted $s_{near}$, and proceed as follows:
	\begin{itemize}
		\item If $|s_{near} - x_{new}| > \delta_{s}$, then add a new witness $s_{new} = x_{new}$ to $S$ and set its representative as $x_{new}$. Add a vertex $v_{new}$ associated with $x_{new}$ to $V_{active}$ and an edge $(v_{cur}, v_{new})$ associated with $\psi_{new}$ to $E$. Then, go to \textbf{Step 3}.
		\item If $|s_{near} - x_{new}| \leq \delta_{s}$, 
		\begin{itemize}
			\item if $\overline{c}_{s_{near}.rep} > c_{new}$, add a vertex $v_{new}$ associated with $x_{new}$ to $V_{active}$ and an edge $(v_{cur}, v_{new})$ associated with $\psi_{new}$ to $E$. Then, update the representative of $s_{near}$ by the newly added vertex, i.e., $s_{near}.rep \leftarrow v_{new}$, and prune the vertex, say, $v_{pre\_near}$ which is previously witnessed by $s_{near}$. If $v_{pre\_near}$ is an inactive vertex, i.e., is such as $v_{a}$ described above, then add $v_{pre\_near}$ to $V_{inactive}$. Otherwise, remove $v_{pre\_near}$ and all its child vertices from the search tree. Then, go to \textbf{Step 3}.
			\item if $\overline{c}_{s_{near}.rep} \leq c_{new}$, go to \textbf{Step 3} directly.
		\end{itemize}
	\end{itemize}
\end{steps}
\ifbool{conf}{}{The steps above are illustrated in Figure \ref{fig:hysstoverview}. First, the algorithm initializes the search tree and the witness state set from the data in $\mathcal{P}^{*}$ as \textbf{Steps 1 - 2}  show. Then, HySST randomly selects between flow regime and jump regime, followed by choosing a random state, which is denoted $x_{rand}$ and represented by a green dot in Figure \ref{fig:hysstoverview} following \textbf{Steps 3 - 4}. The green circle in Figure \ref{fig:hysstoverview} denotes the random state neighborhood of $x_{rand}$ defined by $\delta_{BN}$ in \textbf{Step 5}. Then, among all the active vertices represented by black dots in Figure \ref{fig:hysstoverview}, HySST finds one within the ball of radius $\delta_{BN}$ associated to lowest cost, denoted $v_{cur}$ as is described in \textbf{Step 5}. The inactive vertices represented by blue dots are ignored at this step. Next, HySST propagates forward by applying a randomly selected input to generate $v_{new}$, shown in Figure \ref{fig:hysstoverview}, as is described in \textbf{Step 6}. Note that in Figure \ref{fig:hysstoverview}, $v_{new}$ is not in the closest witness neighborhood of any existing witness, represented by the red circles. In this case, $v_{new}$ is added to $V_{active}$, no vertices need to be pruned, and, therefore, $V_{inactive}$ is not updated.  A witness that equals $\overline{x}_{v_{new}}$ and whose representative is $v_{new}$ is added to the witness state set as is presented in \textbf{Step 7}.}

%% file: sections/algorithm/algorithm.tex
\subsection{HySST Algorithm}
\label{section:hybridRRTframewrok}
Following the overview above, the proposed algorithm is given in Algorithm \ref{algo:hybridRRT}. The inputs of Algorithm \ref{algo:hybridRRT} are the problem $\mathcal{P}^{*} = (X_{0}, X_{f}, X_{u}, (C, f, D, g), c)$, the input library  $ (\mathcal{U}_{C}, \mathcal{U}_{D})$, a parameter $p_{n}\in (0, 1)$, which tunes the probability of proceeding with the flow regime or the jump regime, an upper bound $K\in \mathbb{N}_{>0}$ for the number of iterations to execute, and two tunable sets $X_{c}\supset \overline{C'}$ and $X_{d}\supset D'$, which act as constraints in finding a closest vertex to $x_{rand}$. In addition, HySST requires additional parameters $\delta_{BN}$ and $\delta_{s}$ to tune the radius of random state neighborhood and closest witness neighborhood, respectively. 
\begin{algorithm}[htbp]
	\caption{HySST algorithm}
	\label{algo:hybridRRT}
	\hspace*{\algorithmicindent} \textbf{Input: $X_{0}, X_{f}, X_{u}, \mathcal{H} = (C, f, D, g), (\mathcal{U}_{C}, \mathcal{U}_{D}), p_{n} \in (0, 1)$, $K\in \mathbb{N}_{>0}$, $X_{c}$, $X_{d}$, $\delta_{BN}$ and $\delta_{s}$}
	\footnotesize
	\begin{algorithmic}[1]
		\State $\mathcal{T}.init(X_{0})$;
		\State $V_{active} \leftarrow V$, $V_{inactive} \leftarrow \emptyset$, $S\leftarrow \emptyset$;
		\ForAll{$v_{0}\in V$}
			\If{$is\_vertex\_locally\_the\_best(\overline{x}_{v_{0}}, 0, S, \delta_{s})$}
				\State $(S,$ $V_{active},$ $V_{inactive},$$ E)\leftarrow prune\_dominated\_vertices(v_{0}, S,  V_{active}, V_{inactive}, E)$
			\EndIf
		\EndFor
		\For{$k = 1$ to $K$}
		\State randomly select a real number $r$ from $[0, 1]$;
		\If{$r\leq p_{n}$}
		\State $x_{rand}\leftarrow random\_state(\overline{C'})$;
		\State $v_{cur}$$\leftarrow$ $best\_near\_selection$$(x_{rand}, \;$$V_{active},$$ \delta_{BN},$$ X_{c})$;
		\Else
		\State $x_{rand}\leftarrow random\_state(D')$;
		\State $v_{cur}\leftarrow best\_near\_selection(x_{rand}, V_{active}, \delta_{BN}, X_{d})$;
		\EndIf
		\State $(is\_a\_new\_vertex\_generated, x_{new}, \psi_{new}, cost_{new})\leftarrow new\_state(v_{cur}, (\mathcal{U}_{C}, \mathcal{U}_{D}), \mathcal{H}, X_{u})$
		\If {$is\_a\_new\_vertex\_generated$ \& $is\_vertex\_locally\_the\_best(x_{new}, cost_{new}, S, \delta_{s})$}
		\State $v_{new} \leftarrow V_{active}.add\_vertex(x_{new}, cost_{new})$;
		\State $E.add\_edge(v_{cur}, v_{new}, \psi_{new})$;
		\State $(S, V_{active}, V_{inactive}, E)\leftarrow prune\_dominated\_vertices(v_{new}, S, V_{active}, V_{inactive}, E)$;
		\EndIf
		%		\State \htc{$extend(\mathcal{T}, x_{rand}, (\mathcal{U}_{C}, \mathcal{U}_{D}), \mathcal{H}, X_{u})$;}
		\EndFor
		\State \Return $\mathcal{T}$;
	\end{algorithmic}
\end{algorithm}
Each function in Algorithm \ref{algo:hybridRRT} is defined next.
\subsubsection{$\mathcal{T}.init(X_{0})$} 
The function call $\mathcal{T}.init$ is used to initialize a search tree $\mathcal{T} = (V, E)$.  It randomly selects a finite number of points from $X_{0}$. For each sampling point $x_{0}$, a vertex $v_{0}$ associated with $x_{0}$ is added to $V$. At this step, no edge is added to $E$.
\subsubsection{$return\leftarrow is\_vertex\_locally\_the\_best(x, cost, S, \delta_{s})$}
The function call $is\_vertex\_locally\_the\_best$ describes the conditions under which the state $x$ is considered for addition to the search tree as is shown in Algorithm \ref{algo:isvertexlocallybest}. First, Algorithm \ref{algo:isvertexlocallybest} looks for the closest witness $s_{new}$ to $x$ from the witness set $S$ (line 1). If the closest witness is more than $\delta_{s}$ from $x$, then a new witness is added to $S$ (lines 2 - 6). If $s_{new}$ is just added as a witness or $cost$ is less than the cost of the closest witness's representative (line 7), then the state $x$ with the cost $cost$ is locally optimal and a $true$ signal is returned (line 8). Otherwise, a $false$ signal is returned.
\begin{algorithm}[htbp]
	\caption{$is\_vertex\_locally\_the\_best(x, cost, S, \delta_{s})$}
	\label{algo:isvertexlocallybest}
	\footnotesize
	\begin{algorithmic}[1]
		\State $s_{new}\leftarrow nearest(S, x)$;
		\If{$|x - s_{new}| > \delta_{s}$}
			\State $s_{new}\leftarrow x$
			\State $s_{new}.rep \leftarrow NULL$
			\State $S\leftarrow S\cup \{s_{new}\}$;
		\EndIf
		\If{$s_{new}.rep == NULL$ or $cost < \overline{c}_{s_{new}.rep}$}
			\State \Return true;
		\EndIf
		\State \Return false;
	\end{algorithmic}
\end{algorithm}
\subsubsection{$(S, V_{active}, V_{inactive}, E)\leftarrow prune\_dominated\_vertices(v, S, V_{active}, V_{inactive}, E)$}
Algorithm \ref{algo:prunedominatedvertex} describes the pruning process of dominated vertices. First, Algorithm \ref{algo:prunedominatedvertex} looks for the witness $s_{new}$ that is closest to $\StateofVertex{v}$ and its representative $v_{peer}$ (lines 1 - 2). The previous representative, which is dominated by $v$ in terms of cost, is removed from the active set of vertices $V_{active}$ and is added to the inactive vertices set $V_{inactive}$ (lines 4 - 5). Then, $v$ replaces $v_{peer}$ as the representative of $s_{new}$ (line 7). If $v_{peer}$ is a leaf vertex, then it can also safely be removed from the search tree (lines 8 - 13). The removal of $v_{peer}$ may cause a cascading effect for its parents, if they have already been in the inactive set $V_{inactive}$ and the only reason they were maintained in the search tree was because they were leading to $v_{peer}$. Here, the function call $isleaf(v_{peer})$ returns $true$ signal if $v_{peer}$ is a leaf vertex, which means $v_{peer}$ does not have child vertices (line 8). The function call $parent(v_{peer})$ returns the parent vertex of $v_{peer}$ (line 9). 
\begin{algorithm}[htbp]
	\caption{$(S, V_{active}, V_{inactive}, E)\leftarrow prune\_dominated\_vertices(v, S, V_{active}, V_{inactive}, E)$}
	\label{algo:prunedominatedvertex}
	\footnotesize
	\begin{algorithmic}[1]
		\State $s_{new}\leftarrow nearest(S, \StateofVertex{v})$;
		\State $v_{peer} \leftarrow s_{new}.rep$;
		\If{$v_{peer}!= NULL$}
			\State $V_{active} \leftarrow V_{active}\backslash \{v_{peer}\}$;
			\State $V_{inactive}\leftarrow V_{inactive}\cup \{v_{peer}\}$;
		\EndIf
		\State $s_{new}.rep\leftarrow v$;
		\While{$isleaf(v_{peer})$ and $v_{peer}\in V_{inactive}$}
			\State $v_{parent}\leftarrow parent(v_{peer})$;
			\State $E\leftarrow E\backslash \{(v_{parent}, v_{peer})\}$;
			\State $V_{inactive}\leftarrow V_{inactive}\backslash{v_{peer}}$;
			\State $v_{peer}\leftarrow v_{parent}$;
		\EndWhile
	\end{algorithmic}
\end{algorithm}
\subsubsection{$x_{rand}$$\leftarrow$$random\_state(S)$} 
The function call $random\_state$ randomly selects a point from the set $S\subset \mathbb{R}^{n}$. It is designed to select from $\overline{C'}$ and $D'$ separately depending on the value of $r$ rather than to select from $\overline{C'}\cup D'$. The reason is that if $\overline{C'}$ ($D'$) has zero measure while $D'$ ($\overline{C'}$) does not, the probability that the point selected from $\overline{C'}\cup D'$ lies in $\overline{C'}$ ($D'$, respectively) is zero, which would prevent finding a solution when one exists.
%The selected state is used as a target state in the forthcoming function $extend$. The selection is made in $\overline{C'}\cup g(D)$ because a state can be reached during flow if it is in $\overline{C'}$ and at a jump if it is in $g(D)$.

%	It is straightforward to implement the sampling in $C'$. Note that the explicit expression of $g(D)$ is not always available. In this case, the sampling in $G(D)$ can not be implemented directly. One option is to sample a pair $(x, u)\in D$ and return $g(x, u)$. Then the sampling in $\overline{C'}\cup G(D)$ is implemented by sampling a state $x_{rand}$ in $\overline{C'}\cup D'$. In this case, the function $RANDOM\_STATE$ considers the following cases over the sample $x_{rand}$:
%	\begin{enumerate}
%		\item If $x_{rand}\in \overline{C'}$, then the algorithm returns $x_{rand}$.
%		\item If $x_{rand}\in D'$, then the algorithm samples a $u_{rand}\in U_{D}$ such that $(x_{rand}, u_{rand})\in D$ and returns $g(x_{rand}, u_{rand})$. If $g$ is a set-valued map, then the algorithm returns a sample of $g(x_{rand}, u_{rand})$.
%		\item If $x_{rand}\in \overline{C'}\cap D'$, the algorithm executes a simple random sampling from the set $\{0, 1\}$. If it draws $0$, then go to item (a). Otherwise, go to item (b).
%	\end{enumerate}
\subsubsection{$v_{cur}\leftarrow best\_near\_selection(x_{rand}, V_{active}, \delta_{BN}, X_{\star})$} 
The function call $best\_near\_selection$ searches for a vertex $v_{cur}$ in the active vertex set $V_{active}$ such that its associated state value is in the intersection between the set $X_{\star}$ and $\Ballof{\delta_{BN}}{x_{rand}}$, and has minimal cost where $\star$ is either $c$ or $d$. This function is implemented by solving the following optimization problem.
	\begin{problem}
		\label{problem:nearestneighbor}
		Given $x_{rand}\in \Reals[n]$, a radius $\delta_{BN} > 0$ of the random state neighborhood, a tunable state constraint set $X_{\star}$, and an active vertex set $V_{active}$, solve
		$$
		\begin{aligned}
		\argmin_{v\in V_{active}}& \quad \overline{c}_{v}\\
		\textrm{s.t.}& \quad|\overline{x}_{v} - x_{rand}| \leq \delta_{BN}\\
		& \quad\overline{x}_{v}\in X_{\star}.
		\end{aligned}
		$$
	\end{problem}
The data of Problem \ref{problem:nearestneighbor} comes from the arguments of the $best\_near\_selection$ function call. This optimization problem can be solved by traversing all the vertices in $V_{active}$. %If no solution to Problem \ref{problem:nearestneighborflow}/\ref{problem:nearestneighborjump} is found, then the function call returns the nearest vertex to $x_{rand}$ regardless of the the constraints.

\subsubsection{$(is\_a\_new\_vertex\_generated, x_{new}, \psi_{new}, cost_{new})\leftarrow new\_state(v_{cur}, (\mathcal{U}_{C}, \mathcal{U}_{D}) , \mathcal{H} , X_{u})$} 
\label{section:newstate}
If $\overline{x}_{v_{cur}}\in \overline{C'}\backslash D'$ ($\overline{x}_{v_{cur}}$$\in$$D'\backslash \overline{C'}$), the function call $new\_state$ generates a new solution pair $\psi_{new}$ to hybrid system $\mathcal{H}$ starting from $\overline{x}_{v_{cur}}$ by applying a input signal $\tilde{u}$ (an input value $u_{D}$) randomly selected from $\mathcal{U}_{C}$ ($\mathcal{U}_{D}$, respectively). 
%The solution pair $\psi_{new}$ is computed by simulators of continuous dynamics or discrete dynamics. 
If $\overline{x}_{v_{cur}}$$\in\overline{C'}\cap D'$, then this function generates $\psi_{new}$ by randomly selecting flows or jump. The final state of $\psi_{new} = (\phi_{new}, u_{new})$ is denoted as $x_{new}$. The cost $cost_{new}$ at $x_{new}$ is computed by $cost_{new}\leftarrow\overline{c}_{v_{cur}} + c(\phi_{new})$.
After $\psi_{new}$ and $x_{new}$ are generated, the function $new\_state$ checks if there exists $(t, j)\in \dom \psi_{new}$ such that $\psi_{new}(t, j)\in X_{u}$. If so, then $\psi_{new}$ intersects with the unsafe set and $is\_a\_new\_vertex\_generated\leftarrow false$. Otherwise, this function returns $is\_a\_new\_vertex\_generated\leftarrow true$. 
%\subsubsection{$return \leftarrow is\_vertex\_locally\_the\_best(v_{cur}, x_{new}, \psi_{new}, V, \delta_{s})$}
%The function call $is\_vertex\_locally\_the\_best$ checks if there exists a neighbor vertex of $x_{new}$ in the search that has a less cost. If so, return false. Otherwise, return true. In other words, if the set $\{v\in V: \overline{c}_{v} \leq \overline{c}_{v_{cur}} + c(\psi_{new}), |\overline{x}_{v} - x_{new}| \leq \delta_{s}\}$ is not an empty set, then return false; otherwise, return true.
\subsubsection{$v_{new} \leftarrow V_{active}.add\_vertex(x_{new}, cost_{new})$ and  $E.add\_edge$$(v_{cur}$, $v_{new}$, $\psi_{new})$} 
The function call $V_{active}.add\_vertex(x_{new}, cost_{new})$ adds a new vertex $v_{new}$ to $V_{active}$ such that $\overline{x}_{v_{new}} \leftarrow x_{new}$ and $\overline{c}_{v_{new}} \leftarrow cost_{new}$ and returns $v_{new}$. The function call $E.add\_edge(v_{cur}, v_{new}, \psi_{new})$ adds a new edge $e_{new} = (v_{cur}, v_{new})$ associated with $\psi_{new}$  to $E$. 
\subsection{Solution Checking during HySST Construction}
\label{section:checksolution}
At each iteration, when a new vertex and a new edge are added to the search tree, i.e., $is\_a\_new\_vertex\_generated == true$, a solution checking function is employed to check if a path in $\mathcal{T}$ can be used to construct a motion plan to the given motion planning problem. If this function finds a path
$
p = ((v_{0}, v_{1}), (v_{1}, v_{2}), ..., (v_{n - 1}, v_{n})) = : (e_{0}, e_{1}, ..., e_{n - 1})
$
in $\mathcal{T}$ such that 
\ifbool{conf}{ 1) $\overline{x}_{v_{0}} \in X_{0}$ and  2) $\overline{x}_{v_{n}} \in X_{f}$,}{
	\begin{enumerate}[label=\arabic*)]
		\item $\overline{x}_{v_{0}} \in X_{0}$,
		\item $\overline{x}_{v_{n}} \in X_{f}$,
		\item for each pair of adjacent edges $e_{i}$ and $e_{i + 1}$, where $i = 0, 1, ..., n - 2$, if $\overline{\psi}_{e_{i}}$ and $\overline{\psi}_{e_{i + 1}}$ are both purely continuous, then $\overline{\psi}_{e_{i + 1}}(0, 0)\in C$,
\end{enumerate}}
then the solution pair $\tilde{\psi}_{p}$, defined in (\ref{equation:concatenationpath}), is a motion plan to the given motion planning problem.
%	If such a path is found, then $check\_solution$ assigns $\tilde{\psi}_{p}$ to $\psi_{sol}$ and returns $true$. Otherwise, this function returns $false$. 
\ifbool{conf}{
%	In practice, item 2) is too restrictive. Given $\epsilon > 0$ representing the tolerance with this condition, we implement item 2) as
%	$
%	\text{dist}(\overline{x}_{v_{n}}, X_{f}) \leq \epsilon.
%	$
}{While item 2) above requires that $\overline{x}_{v_{n}}$ belongs to $X_{f}$, in practice, given $\epsilon > 0$ representing the tolerance with this condition, we implement item 2) as
	\begin{equation}
	\label{equation:tolerance}
	\text{dist}(\overline{x}_{v_{n}}, X_{f}) \leq \epsilon.
	\end{equation}}

%% file: sections/mainresults/main.tex
This section analyzes the asymptotic optimality property of HySST algorithm.
\input{sections/mainresults/assumption_costfunctional.tex}
%\input{sections/mainresults/assumption_dynamics.tex}
\input{sections/mainresults/definition_clearance.tex}
\input{sections/mainresults/definition_inflation.tex}

\input{sections/mainresults/assumption_parameters.tex}
\input{sections/mainresults/assumption_input.tex}
\input{sections/mainresults/assumption_sample.tex}
\input{sections/mainresults/assumption_flowcontinuous.tex}
\ifbool{conf}{}{The forthcoming Lemma \ref{lemma:pccontinuouslowerbound} characterizes the probability that the simulated solution pair computed by the function call $new\_state$ in the flow regime is close to the motion plan. 
\input{sections/mainresults/lemma_flow_lemma.tex}}

\input{sections/mainresults/assumption_jumpcontinuous.tex}
\ifbool{conf}{}{The forthcoming Lemma \ref{lemma:pcdiscretelowerbound} characterizes the probability that the simulated solution pair computed by the function call $new\_state$ in the jump regime is close to the motion plan. 
	\input{sections/mainresults/lemma_jump_lemma.tex}}

\ifbool{conf}{}{\input{sections/mainresults/lemma_goodvertexexists.tex}
	\input{sections/mainresults/lemma_selectgoodvertex.tex}
	\input{sections/mainresults/lemma_concatenation_closeness.tex}}

\input{sections/mainresults/definition_robustoptimality.tex}
\input{sections/mainresults/maintheorem.tex}

%% file: sections/mainresults/assumption_costfunctional.tex
The following assumption assumes that the cost functional is Lipchitz continuous along the purely continuous solution pairs, locally bounded at jumps, \pn{and also satisfies} additivity, monotonicity, and non-degeneracy.
\begin{assumption}\label{assumption:costfunctional}
	The cost functional $c: \Sigma_{\phi} \to \Reals_{\geq 0}$ satisfies the following:
	\begin{enumerate}
		\item It is Lipschitz continuous for all continuous solution pairs $(\phi_{0}, u_{0})$ and $(\phi_{1}, u_{1})$ such that $\phi_{0}(0, 0) = \phi_{1}(0, 0)$; specifically,  there exists $K_{c} > 0$ such that $|c(\phi_{0}) - c(\phi_{1})| \leq K_{c}\sup_{ (t, 0)\in \dom \phi_{0}\cap \dom\phi_{1}} \{|\phi_{0}(t, 0) - \phi_{1}(t, 0)|\}$.
		\item For each purely discrete solution pairs $(\phi_{0}, u_{0})$ and $(\phi_{1}, u_{1})$ with one jump such that $\dom \phi_{0} = \dom \phi_{1} = \{0\}\times \{0, 1\}$ and $\phi_{0}(0, 0) = \phi_{1}(0, 0)$, there exists $K_{d} > 0$ such that $ |c(\phi_{0}) - c(\phi_{1})| \leq K_{d}\sup_{ j\in \{0, 1\}} \{|\phi_{0}(0, j) - \phi_{1}(0, j)|\}$.
		\item Consider two solution pair $\psi_{0} = (\phi_{0}, u_{0})$ and $\psi_{1} = (\phi_{1}, u_{1})$ such that their concatenation is $\psi_{0}|\psi_{1}$. The following hold:
		\begin{enumerate}
			\item $c(\phi_{0}|\phi_{1}) = c(\phi_{0}) + c(\phi_{1})$ (additivity);
			\item $c(\phi_{1})\leq c(\phi_{0}|\phi_{1})$ (monotonicity);
			\item For each $t_{2} > t_{1} \geq 0$ such that $(t_{1}, j) \in \dom \psi_{0}$ and $(t_{2}, j) \in \dom \psi_{0}$ for some $j\in \Naturals$, there exists $M_{c} > 0$ such that $t_{2} - t_{1} \leq M_{c} |c(\phi_{0}(t_{2}, j)) - c(\phi_{0}(t_{1}, j))|$ (non-degeneracy during flows).
			\item For each $ j_{1}, j_{2}\in \Naturals$ such that $j_{2} > j_{1}$, $(t,  j_{1}) \in \dom \psi_{0}$ and $(t, j_{2}) \in \dom \psi_{0}$ for some $t\in \Preals$, there exists $M_{d} > 0$ such that $j_{2} - j_{1} \leq M_{d}  |c(\phi_{0}(t, j_{2})) - c(\psi_{0}(t, j_{1}))|$ (non-degeneracy at jumps).
		\end{enumerate} 
	\end{enumerate}  
\end{assumption}
\begin{remark}
\pn{Items 1) and 2) above guarantee that the cost of the nearby solution pairs are bounded by the distance between the solutions. Item 3) above guarantees that the cost of the solution pairs can be computed incrementally and that the global minimum of the cost functional can be found by the optimal motion planning problem.}
\end{remark}
%\pn{Given a state trajectory $\phi\in \Sigma_{\phi}$ such that $(T, J) = \max \dom \phi$, let $\{t_{j}\}_{j = 0}^{J + 1}$ be the sequence such that $\dom \phi = \bigcup_{j = 0}^{J}([t_{j}, t_{j + 1}])\times\{j\}$ where $t_{J + 1} = T$. The cost functional $c:\Sigma_{\phi}\to \Preals$ is formulated in the following form
%\begin{equation}
%	c(\phi) := \left(\sum_{j = 0}^{J}\int_{t_{j}}^{t_{j + 1}}L_{C}(\phi(t, j))\text{d}t\right) + \sum_{j = 0}^{J - 1}L_{D}(\phi(t_{j  + 1}, j))
%\end{equation}
%where $L_{C}: C'\to \Preals$ is called the \emph{flow cost} and $L_{D}: D'\to\Preals$ is called the \emph{jump cost}.}

%% file: sections/mainresults/definition_clearance.tex
Next we define the clearance of the potential motion plans, which is heavily used in the literature; see \cite{kleinbort2018probabilistic}.
\begin{definition}[Safety clearance of a solution pair]
	\label{definition:safetyclearance}
	Given a motion plan $\psi = (\phi, u)$ to the optimal motion planning problem $\mathcal{P}^{*} = (X_{0}, X_{f}, X_{u}, (C, f, D, g), c)$, the safety clearance  $\delta_{s}$ of $\psi = (\phi, u)$ is equal to the maximal $\delta' > 0$ if the following hold:
	\begin{enumerate}[label=\arabic*)]
		\item $\phi(0, 0) + \delta' \mathbb{B} \subset X_{0}$;
		\item  For all $(t, j)\in \dom \psi$, $(\phi(t, j) + \delta'\mathbb{B}, u(t, j) + \delta'\mathbb{B}) \cap X_{u} =\emptyset$.
		\item $\phi(T, J) + \delta' \mathbb{B} \subset X_{f}$, where $(T, J) = \max \dom (\phi, u)$.
	\end{enumerate} 
\end{definition}
\begin{assumption}\label{assumption:clearance}
	The optimal motion plan to the given optimal motion planning problem has positive safety clearance $\delta_{s}$.
\end{assumption}
\begin{definition}[Dynamics clearance of a solution pair]
	\label{definition:dynamicsclearance}
	Given a motion plan $\psi = (\phi, u)$ to the optimal motion planning problem $\mathcal{P}^{*} = (X_{0}, X_{f}, X_{u}, (C, f, D, g), c)$, the dynamics clearance  $\delta_{d}$ of $\psi = (\phi, u)$ is equal to the maximal $\delta' > 0$ \pn{satisfying the following:}
	\begin{enumerate}[label=\arabic*)]
		\item For all $(t, j)\in \dom \psi $ such that $I^{j}$ has nonempty interior, $(\phi(t, j) + \delta'\mathbb{B}, u(t, j) + \delta'\mathbb{B}) \subset C$;
		\item For all $(t, j)\in \dom \psi $ such that $(t, j + 1)\in \dom \psi$, $(\phi(t, j) + \delta'\mathbb{B}, u(t, j) + \delta'\mathbb{B})\subset D$.
	\end{enumerate} 
\end{definition}

Assuming that the optimal motion plan has positive dynamics clearance is restrictive for hybrid systems. Indeed, if the motion plan reaches the boundary of the flow set or of the jump set, then the motion plan has no clearance. To overcome this issue, \ifbool{conf}{the $\delta_{f}$-inflation of hybrid systems, denoted $\mathcal{H}_{\delta_{f}}: = (C_{\delta_{f}}, f_{\delta_{f}}, D_{\delta_{f}}, g_{\delta_{f}})$ for some $\delta_{f} > 0$,  in our previous work is employed to create a positive dynamics clearance; see \cite{wang2022rapidly}.}{the hybrid system $\mathcal{H} = (C, f, D, g)$ is modified as follows.}

%% file: sections/mainresults/definition_inflation.tex
\ifbool{conf}{}{\begin{definition}($\delta_{f}$-inflation of hybrid system)
	\label{definition:inflation}
	Given a hybrid system $\mathcal{H} = (C, f, D, g)$ and $\delta_{f} > 0$, the $\delta_{f}$-inflation of the hybrid system $\mathcal{H}$, denoted $\mathcal{H}_{\delta_{f}}$, is given by
	\begin{equation}
		\mathcal{H}_{\delta_{f}}: \left\{              
		\begin{aligned}               
		\dot{x} & = f_{\delta_{f}}(x, u)     &(x, u)\in C_{\delta_{f}}\\                
		x^{+} & =  g_{\delta_{f}}(x, u)      &(x, u)\in D_{\delta_{f}}\\                
		\end{aligned}   \right. 
		\label{model:inflatedhybridsystem}
		\end{equation}
	 where
	\begin{enumerate}[label=\arabic*)]
		\item $C_{\delta_{f}} := \{(x, u)\in \mathbb{R}^{n}\times \mathbb{R}^{m}: \exists (y,  v)\in C \text{ such that } x\in y + \delta_{f} \mathbb{B}, u\in v+ \delta_{f} \mathbb{B}\}$,
		\item $f_{\delta_{f}}(x, u) := f(x, u)\quad  (x, u)\in C_{\delta_{f}}$,
		\item $D_{\delta_{f}} := \{(x, u)\in \mathbb{R}^{n}\times \mathbb{R}^{m}: \exists (y,  v)\in D \text{ such that } x\in y + \delta_{f} \mathbb{B}, u\in v+ \delta_{f} \mathbb{B}\}$,
		\item $g_{\delta_{f}}(x, u) := g(x, u)\quad  (x, u)\in D_{\delta_{f}}$.
	\end{enumerate}
\end{definition}
Next we show that a motion plan to the original motion planning problem is also a motion plan to the motion planning problem for its $\delta_{f}$-inflation.
\begin{proposition}(Proposition 5.21 in \cite{nwang2021rapid_tr})
	\label{proposition:motioninflated}
	Given a motion planning problem $\mathcal{P}^{*} = (X_{0}, X_{f}, X_{u}, (C, f, D, g), c)$ in Problem \ref{problem:optimalmotionplanning}, if $\psi$ is a motion plan to $\mathcal{P}^{*}$, then $\psi$ is also a motion plan to the motion planning problem $\mathcal{P}_{\delta_{f}}^{*} = (X_{0}, X_{f}, X_{u},  (C_{\delta_{f}}, f_{\delta_{f}}, D_{\delta_{f}}, g_{\delta_{f}}), c)$ with positive dynamics clearance, denoted $\delta_{d}$, such that $\delta_{d} \geq\delta_{f} $, where $(C_{\delta_{f}}, f_{\delta_{f}}, D_{\delta_{f}}, g_{\delta_{f}})$ is the $\delta_{f}$-inflation of hybrid system of $(C, f, D, g)$ for any $\delta_{f} > 0$.
\end{proposition}}With both safety clearance and dynamics clearance defined, we are ready to define the clearance of the solution pair.
\begin{definition}[Clearance of a solution pair]
	\label{definition:clearance}
	Given a motion plan $\psi = (\phi, u)$ to the optimal motion planning problem $\mathcal{P}^{*} = (X_{0}, X_{f}, X_{u}, (C, f, D, g), c)$, the clearance of $\psi = (\phi, u)$, denoted $\delta$, is defined as the minimum of its safety clearance $\delta_{s}$ and dynamics clearance $\delta_{d}$, i.e., $\delta := \min \{\delta_{s}, \delta_{d}\}$.
\end{definition}
\ifbool{conf}{}{Next we show that the existing motion plan with positive safety clearance has positive clearance for the motion planning problem for the $\delta_{f}$-inflation of the original hybrid system.
\begin{lemma}(Lemma 5.22 in \cite{nwang2021rapid_tr})
	\label{lemma:motionplanintlated}
	Let $\psi$ be a motion plan to the motion planning problem $\mathcal{P}^{*} = (X_{0}, X_{f}, X_{u}, (C, f, D, g), c)$ formulated as Problem \ref{problem:optimalmotionplanning} with positive safety clearance $\delta_{s} > 0$. Then\pn{,} for arbitrary $\delta_{f} > 0$, $\psi$ is a motion plan to the motion planning problem $\mathcal{P}^{*}_{\delta_{f}} = (X_{0}, X_{f}, X_{u}, (C_{\delta_{f}}, f_{\delta_{f}}, D_{\delta_{f}}, g_{\delta_{f}}), c)$ with clearance, denoted $\delta$, such that $\delta \geq \min \{\delta_{s}, \delta_{f}\}$, where $\mathcal{H}_{\delta_{f}} = (C_{\delta_{f}}, f_{\delta_{f}}, D_{\delta_{f}}, g_{\delta_{f}})$ is the $\delta_{f}$-inflation of $\mathcal{H} = (C, f, D, g)$.
\end{lemma}}

%% file: sections/mainresults/assumption_parameters.tex
The following assumption relating the clearance $\delta$ of the optimal motion plan with the algorithm parameters $\delta_{BN}$ and $\delta_{s}$ is necessary to establish the optimality property.
\begin{assumption}\label{assumption:parameters}
	Given the clearance $\delta$ of the optimal motion plan, the parameters $\delta_{BN}$ and $\delta_{s}$ need to satisfy the following relationship
	$$
		\delta_{BN} + 2\delta_{s} < \delta.
	$$
\end{assumption}

%% file: sections/mainresults/assumption_input.tex
The following assumption is imposed on the input library.
\begin{assumption}
	\label{assumption:inputlibrary}
	The input library $(\mathcal{U}_{C}, \mathcal{U}_{D})$ is such that
	\begin{enumerate}[label=\arabic*)]
		\item Each input signal in $\mathcal{U}_{C}$ is constant and $\mathcal{U}_{C}$ includes all possible input signals such that their time domains are subsets of the interval $[0, T_{m}]$ for some $T_{m} > 0$ and their images belong to $U_{C}$. In other words, there exists $T_m > 0$ such that $\mathcal{U}_{C} = \{ \tilde{u} : \dom \tilde{u} = [0,T] \subset [0,T_m], \tilde{u} \text{ is } \text{constant} \text{ and } \rge \tilde{u} \in U_C\}$;
		\item $\mathcal{U}_D = U_D$.
	\end{enumerate}
\end{assumption}

%% file: sections/mainresults/assumption_sample.tex
The following assumption is imposed on the random selection in HySST.
\begin{assumption}
		\label{assumption:uniformsample}
%		\begin{enumerate}
The probability distributions of the random selection in the function calls $T.init$, $random\_state$, and  $new\_state$ are the uniform distribution.
%			\item \htc{The probability distribution used for the selection of a state in the function call $random\_state$ is uniform distribution};
%			\item \htc{The probability distribution used for the selection of input signals/values in the function call $new\_state$ is uniform distribution}.
%		\end{enumerate}
	\end{assumption}

%% file: sections/mainresults/assumption_flowcontinuous.tex
The following assumptions are imposed on the flow map $f$ and the jump map $g$ of the hybrid system $\mathcal{H}$ in (\ref{model:generalhybridsystem}).
\begin{assumption}
	\label{assumption:flowlipschitz}
	The flow map $f$ is Lipschitz continuous. In particular, there exist $K^{f}_{x}, K^{f}_{u}\in \mathbb{R}_{>0}$ such that, for all $(x_{0}, x_{1}, u_{0}, u_{1})$ such that $(x_{0}, u_{0}) \in C$, $(x_{0}, u_{1}) \in C$, and $(x_{1}, u_{0}) \in C$,
	$$
	\begin{aligned}
	|f(x_{0}, u_{0}) - f(x_{1}, u_{0})|&\leq K^{f}_{x}|x_{0} - x_{1}|\\
	|f(x_{0}, u_{0}) - f(x_{0}, u_{1})|&\leq K^{f}_{u}|u_{0} - u_{1}|.
	\end{aligned}
	$$
\end{assumption}

%% file: sections/mainresults/lemma_flow_lemma.tex
\begin{lemma} (\cite[Lemma 5.10]{nwang2021rapid_tr})
	\label{lemma:pccontinuouslowerbound}
	Given a hybrid system $\mathcal{H}$ that satisfies Assumption \ref{assumption:flowlipschitz} and an input library that satisfies item 1) of Assumption \ref{assumption:inputlibrary}, let $\psi = (\phi, u)$ be a purely continuous solution pair to $\mathcal{H}$ with clearance $\delta > 0$, $(\tau, 0) = \max \dom \psi$, and constant input function $u$. Suppose that $\tau \leq T_{m}$, where $T_{m}$ comes from item 1) in Assumption \ref{assumption:inputlibrary}. Suppose Assumption \ref{assumption:uniformsample} is satisfied.
	Let $\psi_{new} = (\phi_{new}, u_{new})$ be the purely continuous solution pair generated by the function $new\_state$ in Algorithm \ref{algo:hybridRRT} and initial state $\overline{x}_{v_{cur}} = \phi_{new}(0, 0) \in \phi(0, 0) + \delta\mathbb{B}$.
	Then, for each $\delta_{c} \in (0, \delta]$, there exists $\gamma_{f}\in (0, 1)$ such that 
%	Then, for each $\delta_{c} \in (0, \delta]$, there exists $p_{t}\in (0, 1]$ such that 
	\begin{equation}
	\label{equation:lemmaflow}
	\begin{aligned}
%		&Pr(E_{1} \& E_{2})\geq \\
%		&p_{t}\frac{\zeta_{n} \left(\max \left\{\min \left\{\frac{\delta_{c}}{K^{f}_{u}\tau \exp(K_{x}^{f}\tau)}, \delta\right\}, 0\right\}\right)^{m}}{\mu(U_{C})} =: \gamma_{f}.
Pr(E_{1} \& E_{2})\geq  \gamma_{f}.
	\end{aligned}
	\end{equation}
	where 
	\begin{enumerate}
		\item $E_{1}$ denotes the event that $\phi$ and $\phi_{new}$ are $(\overline{\tau}, \delta)$-close where $(\tau', 0) = \max \dom \phi_{new}$ and $\overline{\tau} = \max (\tau, \tau')$;
		\item $E_{2}$ denotes the event that $x_{new} = \phi_{new}(\tau', 0)\in \phi(\tau, 0) + \delta_{c} \mathbb{B}$ where $x_{new}$ stores the final state of $\phi_{new}$ in the function call $new\_state$ as is introduced in Section \ref{section:newstate},
	\end{enumerate} 
%and $\zeta_{n}$ is given in (\ref{equation:zetan}), $\mu(U_{C})$ denotes the Lebesgue measure of $U_{C}$, and $K_{x}^{f}$ and $K_{u}^{f}$ come from Assumption \ref{assumption:flowlipschitz}.  
\end{lemma}

%% file: sections/mainresults/assumption_jumpcontinuous.tex
\begin{assumption}
	\label{assumption:pcjumpmap}
	The jump map $g$ is such that there exist $K^{g}_{x}\in \mathbb{R}_{>0}$ and $K^{g}_{u}\in \mathbb{R}_{>0}$ such that, for all $(x_{0}, u_{0}) \in D$ and $(x_{1}, u_{1}) \in D$,
	$$
	|g(x_{0}, u_{0}) - g(x_{1}, u_{1})|\leq K^{g}_{x}|x_{0} - x_{1}| + K^{g}_{u}|u_{0} - u_{1}|.
	$$
\end{assumption}

%% file: sections/mainresults/lemma_jump_lemma.tex
\begin{lemma}(\cite[Lemma 5.13]{nwang2021rapid_tr})
	\label{lemma:pcdiscretelowerbound}
	Given a hybrid system $\mathcal{H}$ that satisfies Assumption \ref{assumption:pcjumpmap} and an input library that satisfies item 2) of Assumption \ref{assumption:inputlibrary}, let $\psi = (\phi, u)$ be a purely discrete solution pair to $\mathcal{H}$ with a single jump, i.e., $\max \dom \psi = (0, 1)$ and clearance $\delta > 0$. 
	Let $\psi_{new} = (\phi_{new}, u_{new})$ be the purely discrete solution pair generated by the function $new\_state$ in Algorithm \ref{algo:hybridRRT} with initial state $\overline{x}_{v_{cur}} = \phi_{new}(0, 0)\in \phi(0, 0) + \delta\mathbb{B}$.
	Then, for any $\delta_{c}\in (0, \delta]$,  there exists $\gamma_{g}\in (0, 1)$ such that
	\begin{equation}
	Pr(E)\geq \gamma_{g}
	\end{equation}
	where $E$ denotes the event that $x_{new} = \phi_{new}(0, 1) \in \phi(0, 1) + \delta_{c}\mathbb{B}$, $x_{new}$ stores the final state of $\phi_{new}$ in the function call $new\_state$ as is introduced in Section \ref{section:newstate}.
%	 $\zeta_{n}$ is given in (\ref{equation:zetan}), $\mu(U_{D})$ denotes the Lebesgue measure of $U_{D}$, and $K^{g}_{x}$ and $K^{g}_{u}$ come from Assumption \ref{assumption:pcjumpmap}.  
%	where $E$ denotes the event that $\phi$ and $\phi'$ are $(\tau, \kappa \delta)$-close, $\zeta_{D}$ is the Lebesgue measure of the unit ball in $\mathbb{R}^{m}$, $\mu(U_{C})$ denotes the Lebesgue measure of $U_{C}$, and $K_{x}^{f}$ and $K_{u}^{f}$ are from Assumption \ref{assumption:flowlipschitz}.  
%	
%	
%	Let $\psi = (\phi, u)$ be a purely discrete solution pair with clearance $\delta$ with a single jump, i.e., $\max \dom \psi = (0, 1)$. Denote by $x_{0}, x_{1}$ the state $\phi(0, 0)$, $\phi(0, 1)$ respectively. Suppose that  Assumption \ref{assumption:pcjumpmap}  and the second item in Assumption \ref{assumption:inputlibrary} are satisfied. 
%	Suppose that the $new\_state$ function proceeds with $flag = jump$, begins at $x'_{0}\in x_{0} + \delta\mathbb{B}$ and ends at $x'_{1}$. Then for any $\kappa \in (0, 1]$, we have that:
%	\begin{equation}
%		Pr(x'_{1} \in x_{1} + \kappa \delta\mathbb{B})\geq  \frac{\xi_{D} \left(\max \left(\min \left(\frac{(\kappa - \K^{g}_{x})\delta}{K^{g}_{u}}, \delta\right), 0\right)\right)^{m} }{\mu(U_{D})}
%	\end{equation}
%	where $\xi_{D}$ is the Lebesgue measure of the unit ball in $\mathbb{R}^{m}$, $\mu(U_{D})$ denotes the Lebesgue measure of $U_{D}$, and $K^{g}_{x}$ and $K^{g}_{u}$ are from Assumption \ref{assumption:pcjumpmap}.  
\end{lemma}

%% file: sections/mainresults/lemma_goodvertexexists.tex
The forth coming Lemma \ref{lemma:goodvertexexists} shows that the pruning process guarantees the existence of a vertex close to the optimal motion planning with a decreasing cost if a vertex that is close enough to the optimal motion plan has been generated. 
\begin{lemma}(Lemma 27 in \cite{li2016asymptotically})\label{lemma:goodvertexexists}
	Suppose Assumption \ref{assumption:clearance} and Assumption \ref{assumption:parameters} hold. Let $\delta_{c} = \delta - \delta_{BN} - 2\delta_{s}$. If vertex $v\in V_{active}$ is generated at iteration $k$ so that $\StateofVertex{v}\in \mathcal{B}_{\delta_{c}}(x^{*})$ where $x^{*}$ is a state on the motion plan with positive clearance $\delta$, then for every iteration $k' > k$, then there exists a vertex $v'\in V_{active}$ so that $\StateofVertex{v'}\in \Ballof{(\delta - \delta_{BN})}{x^{*}}$ and $\CostofVertex{v'} \leq \CostofVertex{v}$.
\end{lemma}
%\begin{remark}
%	When a trajectory is generated that ends in $\mathcal{B}_{\delta_{c}}(x^{*})$, Lemma \ref{lemma:goodvertexexists} guarantees that there will always be an active vertex associated with a state in $\mathcal{B}_{\delta}(x^{*})$. 
%\end{remark}

%% file: sections/mainresults/lemma_selectgoodvertex.tex
The forthcoming Lemma \ref{lemma:selectgoodvertex} characterize the probability that a vertex that is close to the optimal motion plan is selected by the $best\_near\_selection$ function call.
\begin{lemma}(Lemma 28 in \cite{li2016asymptotically})\label{lemma:selectgoodvertex}
	Suppose Assumption \ref{assumption:clearance}, Assumption \ref{assumption:inputlibrary}, Assumption \ref{assumption:uniformsample} hold, if there exists $v\in V_{active}$ such that $\StateofVertex{v}\in \Ballof{\delta_{c}}{x^{*}}$ at iteration $k$, then the probability that $best\_near\_selection$ for propagation a vertex $v'$ such that $\StateofVertex{v'}\in \Ballof{\delta}{x^{*}}$ can be lower bounded by a positive constant $\gamma_{select}$ for every $k' > k$.
\end{lemma}

%% file: sections/mainresults/lemma_concatenation_closeness.tex
%\begin{lemma}
%	\label{proposition:concatenateclose}
%	Given compact hybrid arcs $\phi_{1}$,  $\phi_{2}$,  $\phi'_{1}$, and $\phi'_{2}$ such that $\phi_{1}$ and $\phi'_{1}$ are $(\tau_{1}, \epsilon_{1})$-close and $\phi_{2}$ and $\phi'_{2}$ are $(\tau_{2}, \epsilon_{2})$-close, where $(T_{1}, J_{1}) = \max \dom \phi_{1}$, $(T'_{1}, J'_{1}) = \max \dom \phi'_{1}$ and $\tau_{1} = \max \{T_{1} + J_{1}, T'_{1} + J'_{1}\}$, then $\phi_{1}|\phi_{2}$ and $\phi'_{1}|\phi'_{2}$ are $(\tau_{1}+ \tau_{2}, \max\{\epsilon_{1}, \epsilon_{2}\})$-close.
%\end{lemma}

%% file: sections/mainresults/definition_robustoptimality.tex
%\begin{definition}[Asymptotic $\delta$-robust near optimality] Given arbitrary $\delta > 0$, let $c^{*}$ denote the minimum cost over all motion plans to $\mathcal{P} = (X_{0}, X_{f}, X_{u}, (C, f, D, g))$ that has positive clearance $\delta$. Let $Y_{n}^{ALG}$ denote the minimum cost value among all trajectories returned by algorithm $ALG$ at iteration $n$ for the same problem. $ALG$ is asymptotically $\delta$-robust near optimal if for all independent runs: 
%$$
%	Pr\left(\left\{\sup_{n\to \infty} Y_{n}^{ALG} \leq (1 + \alpha \delta)c^{*}\right\}\right) = 1
%$$
%for some constant $\alpha \geq 0$.
%\end{definition}

%% file: sections/mainresults/maintheorem.tex
\ifbool{conf}{}{Note that Lemmas \ref{lemma:selectgoodvertex}, \ref{lemma:pccontinuouslowerbound}, and \ref{lemma:pcdiscretelowerbound} provide lower bounds over the probability of executing correct selection in $best\_near\_selection$ and correct propagation in $new\_state$, and Lemma \ref{lemma:goodvertexexists} guarantees that the pruning process will not prune any vertex that help decrease the cost of the generated motion plan.} Then, we are ready to provide our main result showing that by feeding the the inflation of the original hybrid system, HySST would find a solution such that the cost is close to the minimal cost regardless of the positive dynamics clearance. \ifbool{conf}{See \cite{nwang2023sst} for a detailed proof.}{}
\begin{theorem}\label{theorem:asymptoticoptimality}
	Given an optimal motion planning problem $\mathcal{P}^* = (X_{0}, X_{f}, X_{u}, (C, f, D, g), c)$, suppose Assumptions \ref{assumption:costfunctional},   \ref{assumption:parameters}, \ref{assumption:inputlibrary}, \ref{assumption:uniformsample}, \ref{assumption:flowlipschitz}, and \ref{assumption:pcjumpmap} are satisfied and that there exists an optimal motion plan $\psi^{*} = (\phi^{*}, u^{*})$ to $\mathcal{P}^* $ satisfying Assumption \ref{assumption:clearance} for some $\delta_{s} > 0$. When HySST is used to solve the motion planning problem  $\mathcal{P}^*_{\delta_{f}} = (X_{0}, X_{f}, X_{u}, (C_{\delta_{f}}, f_{\delta_{f}}, D_{\delta_{f}}, g_{\delta_{f}}), c)$ where, for some $\delta_{f}> 0$, $(C_{\delta_{f}}, f_{\delta_{f}}, D_{\delta_{f}}, g_{\delta_{f}})$ denotes $\delta_{f}$-inflation of $(C, f, D, g)$, the probability that HySST finds a motion plan $\psi = (\phi, u)$ such that
	$
	c(\phi) \leq (1 + \alpha \delta)c(\phi^*)
	$ \pn{converges to one} as the number of \pn{iterations} $k$ goes to infinity for some constant $\pn{\alpha}\geq 0$, where $\delta = \min \{\delta_{s}, \delta_{f}\}$.
\end{theorem}
	\ifbool{conf}{}{\begin{proof}
			A sketch of the proof is given as follows. First, because the cost function $c$ is Lipchitz continuous, non-decreasing, and non-degenerate (Items 1, 2, 3.c, and 3.d in Assumption \ref{assumption:costfunctional}), and Lemma \ref{lemma:motionplanintlated} guarantees that the motion plan has positive clearance, a sequence of balls $\{\mathcal{B}_{i}\}_{i = 0}^{N}$ centered at the motion plan is constructed such that the costs of the truncation between the centers of the consecutive balls are constant. Then, Lemma \ref{lemma:selectgoodvertex} shows that the probability that $best\_near\_selection$ selects a vertex in $\mathcal{B}_{i}$ is positive and Lemma \ref{lemma:pccontinuouslowerbound} and Lemma \ref{lemma:pcdiscretelowerbound} show that the probability that $new\_state$ generates a new vertex in $\mathcal{B}_{i + 1}$ is positive. Therefore, by using induction, we can show that the probability that HySST generates a solution that is close to the optimal motion plan converges to one as the number of iterations goes to infinity. Lemma \ref{lemma:goodvertexexists} guarantees that the pruning process will only improve the quality of the generated motion plan. Items 1) - 2) in Assumption \ref{assumption:costfunctional} guarantee that the closeness between the generated motion plan and the optimal motion plan leads to a bounded differences between their costs.
		
		Since $\psi^*$ is assumed to have positive safety clearance $\delta_{s} > 0$ and HySST is used to solve $\mathcal{P}^*_{\delta_{f}}$, then according to Lemma \ref{lemma:motionplanintlated}, $\psi^*$ is a motion plan to $\mathcal{P}^*_{\delta_{f}}$ with a positive clearance at least $\delta = \min \{\delta_{s}, \delta_{f}\}$. 
	
	 Then consider a sequence of hybrid time instance $S_{hti}:=\{(T_{i}, J_{i})\in \dom \psi^{*}\}_{i = 0}^{N}$ such that $0 = T_{0} + J_{0} < T_{1} + J_{1} <...< T_{N} + J_{N} = T + J$ where $(T, J) = \max\dom \psi^*$, and, for $i = 0, 1, ..., N - 1$ and some constant $\Delta c_{f}\in\Preals$, either of the following holds
	\begin{enumerate}
		\item $J_{i} = J_{i + 1} $ and $c(\Truncation{\phi^*}{T_{i}, J_{i}}{T_{i + 1}, J_{i + 1}}) = \Delta c_{f}$
		\item $J_{i + 1} = J_{i}  + 1$ and $T_{i + 1} = T_{i}$.
	\end{enumerate}
Namely, using such a sequence of hybrid time instance, $\psi^*$ can be truncated into either, say, $N_{f}$ number of purely continuous segments whose cost equals $\Delta c_{f}$, or, say, $N_{g}$ number of purely discrete segments with a single jump such that \begin{equation}\label{equation:nfng}
	N_{f} + N_{g} = N.
\end{equation} 
Denote the minimal cost of the purely discrete segments as $\Delta c_{j}$.

Constructing the sequence following the second item above is trivial because it can be implemented by selecting all the hybrid time instances before and after all the jumps in $\psi^*$. Note that the cost  function is continuous, non-decreasing and non-degeneracy along the state trajectory during flows because 3.a - 3.c in Assumption \ref{assumption:costfunctional} are assumed. Suppose that $[T_{i} , T_{i} + \Delta T]\times \{J_{i}\}\subset\dom \psi^*$ for some $\Delta T > 0$. Then there exists $\Delta c_{f} > 0$ such that $0 = c(\Truncation{\phi}{T_{i}, J_{i}}{T_{i} + 0,  J_{i}}) <\Delta c_{f} \leq c(\Truncation{\phi}{T_{i}, J_{i}}{T_{i} + \Delta T,  J_{i}})$. In the meantime, given this $\Delta c_{f}$, there exists $\Delta t \in (0, \Delta T]$ such that $c(\Truncation{\phi}{T_{i}, J_{i}}{T_{i} + \Delta t, J_{i}}) = \Delta c_{f}$. Therefore, the sequence of the hybrid time instances following the first item above exists for a sufficiently small $\Delta c_{f}$.

From $S_{hti}$, we can build a sequence of balls $S_{b} := \{\Ballof{r_{i}}{c_{i}}\}_{i = 0}^{N}$ in $\mathbb{R}^{n}$ such that
\begin{enumerate}
	\item $c_{i} := \phi^*(T_{i}, J_{i})$ for all $i = 0, 1, ..., N$.
	\item $r_{i} := \delta$.
\end{enumerate}

Then, denote $\delta_{c} = \delta - \delta_{BN} - 2\delta_{c}$. Let $A^{(k)}_{i}$ denote event that at the $k$-th iteration of HySST, a solution pair $\psi_{new}  = (\phi_{new}, u_{new})$ is generated such that 
\begin{enumerate}
	\item $\phi_{new}(0, 0)\in \phi^*(T_{i}, J_{i}) + \delta \mathbb{B}$;
	\item $\phi_{new}(T_{new}, J_{new})\in \phi^*(T_{i + 1}, J_{i + 1}) + \delta_{c} \mathbb{B}$ where $(T_{new}, J_{new}) = \max \dom \phi_{new}$;
	\item $\phi_{new}$ is $(\overline{\tau}, \delta)$-close to $\Truncation{\phi^*}{T_{i}, J_{i}}{T_{i + 1}, J_{i + 1}}$, where $(T_{new}, J_{new}) = \max \dom \phi_{new}$, $\overline{\tau} = \max \{T_{new} + J_{new}, T_{i + 1} - T_{i} + J_{i + 1} - J_{i}\}$.
\end{enumerate}
Then, let $E^{(k)}_{i}$ denote the event that from iteration $j$ from $1$ to $k$, at least one such $\psi_{new}$ is generated, namely, at least one $j$ such that $A_{i}^{(j)}$ occurs.

Based on the definitions of the events $E^{(k)}_{i}$ and $A^{(k)}_{i}$, the probability that event $E^{(k)}_{i}$ fails, denoted $\neg E^{(k)}_{i}$, depends on a sequence of $A_{i}$ events failing:
\begin{equation}\label{equation:gammaa}
	Pr(\neg E^{(k)}_{i}) = Pr(\neg A^{(1)}_{i}) \cdot Pr(\neg A^{(2)}_{i} | \neg A^{(1)}_{i}) \cdots Pr(\neg A^{(k)}_{i} | \bigcap_{j = 1}^{k - 1} \neg A^{(j)}_{i})
\end{equation}
The probability that $\neg A^{k}_{i}$ occurs given $\bigcap_{j = 1}^{k - 1} \neg A^{(j)}_{i}$ is equivalent with 
\begin{enumerate}
	\item the probability that HySST fails to generate a vertex in the search tree such that its associated state is in $\phi^{*}(T_{i}, J_{i}) + \delta_{c}\mathbb{B}$, plus
	\item  the probability that HySST generates a vertex in the search tree such that its associated state is in $\phi^{*}(T_{i}, J_{i}) + \delta_{c}\mathbb{B}$ but fails to generate a new solution pairs $\psi_{new} = (\phi_{new}, u_{new})$ such that $\phi_{new}(T_{new}, J_{new}) \in \phi^{*}(T_{i + 1}, J_{i + 1}) + \delta_{c}\mathbb{B}$ where $(T_{new}, J_{new}) = \max \dom \phi_{new}$, which can be characterized by the $\gamma_{select}$ in Lemma \ref{lemma:selectgoodvertex}, and $\gamma_{fg} := \min \{\gamma_{f}, \gamma_{g}\}$ where $\gamma_{f}$ comes from Lemma \ref{lemma:pccontinuouslowerbound} and $\gamma_{g}$ comes from Lemma \ref{lemma:pcdiscretelowerbound}.
\end{enumerate}
Therefore, 
\begin{equation}\label{equation:prgamma}
\begin{aligned}
Pr(\neg A^{(k)}_{i} | \bigcap_{j = 1}^{k - 1} \neg A^{(j)}_{i}) &= Pr(\neg E^{k}_{i - 1}) + Pr(E^{k}_{i - 1})\cdot Pr(\phi_{new}(T_{new}, J_{new}) \notin \phi^{*}(T_{i}, J_{i}) + \delta_{c}\mathbb{B})\\
&\leq Pr(\neg E^{k}_{i - 1}) + Pr(E^{k}_{i - 1})\cdot (1 - \gamma_{select}\gamma_{fg})\\
&= 1 - Pr(E^{k}_{i - 1})\gamma_{select}\gamma_{fg})
\end{aligned}
\end{equation}
Using (\ref{equation:gammaa}) and (\ref{equation:prgamma}), we have
\begin{equation}\label{equation:Eki}
Pr(E^{(k)}_{i}) \geq 1 - \prod_{j = 1}^{k} (1 - Pr(E^{j}_{i - 1})\gamma_{select}\gamma_{fg}))
\end{equation}

Then we can use the following induction to prove that if  $\lim_{k\to\infty} Pr(E^{(k)}_{i})  = 1$, then $\lim_{k\to\infty} Pr(E^{(k)}_{i + 1})  = 1$. If this holds, then eventually, HySST will eventually generate a solution that is close to the optimal motion plan.

\textbf{For the base case,} $\lim_{k\to \infty}Pr(E^{k}_{0}) = 1$ is true, because when the $random\_state$ is executed for infinite number of times, eventually a sample will fall into $\phi^{*}(0, 0) + \delta_{c}\mathbb{B}$. Then we have
\begin{equation}
	Pr(E^{k}_{1}) \geq 1 - (1 - \gamma_{select}\gamma_{fg})^{k}.
\end{equation} which leads to $\lim_{k\to \infty} Pr(E^{k}_{1})  = 1$.

\textbf{For the induction step,} assuming that $\lim_{k\to \infty}Pr(E^{k}_{i}) = 1$, we need to show that $\lim_{k\to \infty}Pr(E^{k}_{i + 1}) = 1$.
Define $y_{i}^{(k)} = \prod_{j = 1}^{k}(1 - Pr(E_{i - 1}^{j})\gamma_{select}\gamma_{fg})$. The logarithm of $y_{i}^{(k)}$ behaves like the follows:
\begin{equation}
	\log y_{i}^{(k)} = \log \prod_{j = 1}^{k}(1 - Pr(E_{i - 1}^{j})\gamma_{select}\gamma_{fg}) = \sum_{j = 1}^{k} \log (1 - Pr(E_{i - 1}^{j})\gamma_{select}\gamma_{fg}).
\end{equation}
The above leads to the following:
\begin{equation}
	\log y_{i}^{(k)} < \sum_{j = 1}^{k}  - Pr(E_{i - 1}^{j})\gamma_{select}\gamma_{fg} = -\gamma_{select}\gamma_{fg} \sum_{j = 1}^{k}Pr(E_{i - 1}^{j})
\end{equation}

Note that according to the induction assumption, $\lim_{k\to \infty}Pr(E^{k}_{i}) = 1$, then $\lim_{k\to \infty} \sum_{j = 1}^{k}Pr(E_{i}^{j}) = \infty$. Therefore, $$\lim_{k \to\infty}\log y_{i + 1}^{(k)}  =-\gamma_{select}\gamma_{fg} \lim_{k \to\infty}\sum_{j = 1}^{k}Pr(E_{i}^{j}) = -\infty$$
which implies that 
$$\lim_{k \to\infty} y_{i + 1}^{(k)}  = 0.$$

With (\ref{equation:Eki}) and $\lim_{k \to\infty} y_{i + 1}^{(k)}  = 0$, it can be shown that
$$
	\lim_{k\to \infty}Pr(E^{k}_{i + 1}) = 1 - \lim_{k \to\infty} y_{i + 1}^{(k)} = 1 - 0 =1.
$$

Therefore, the probability that HySST finds a motion plan that is close to $\psi^*$ is converging to $1$ as the number of iteration goes to infinite.

Note that when the event $A_{i}^{k}$ occurs, $\phi_{new}$ is $(\overline{\tau}, \delta)$-close to $\Truncation{\phi^*}{T_{i}, J_{i}}{T_{i + 1}, J_{i + 1}}$, and $\phi_{new}$ and $\Truncation{\phi^*}{T_{i}, J_{i}}{T_{i + 1}, J_{i + 1}}$ are either both purely continuous or purely discrete. If $\phi_{new}$ and $\Truncation{\phi^*}{T_{i}, J_{i}}{T_{i + 1}, J_{i + 1}}$ are both purely continuous, since item 1 in Assumption \ref{assumption:costfunctional} is assumed, then
\begin{equation}
\label{equation:flowopt}
	\begin{aligned}
	|c(\phi_{new}) - c(\Truncation{\phi^*}{T_{i}, J_{i}}{T_{i + 1}, J_{i + 1}})| &\leq K_{c}\sup_{\forall (t, 0)\in \dom \phi_{new}\cap \dom\Truncation{\phi^*}{T_{i}, J_{i}}{T_{i + 1}, J_{i + 1}}} \{|\phi_{new}(t, 0) - \Truncation{\phi^*}{T_{i}, J_{i}}{T_{i + 1}, J_{i + 1}}(t, 0)|\}\\
	& = K_{c}\delta
	\end{aligned}
\end{equation}
If $\phi_{new}$ and $\Truncation{\phi^*}{T_{i}, J_{i}}{T_{i + 1}, J_{i + 1}}$ are both purely discrete, since item 2 in Assumption \ref{assumption:costfunctional} is assumed, then
\begin{equation}\label{equation:jumpopt}
	\begin{aligned}
	|c(\phi_{new}) - c(\Truncation{\phi^*}{T_{i}, J_{i}}{T_{i + 1}, J_{i + 1}})| \leq K_{d}\sup_{\forall j\in \{0, 1\}} \{|\phi_{new}(0, j) - \Truncation{\phi^*}{T_{i}, J_{i}}{T_{i + 1}, J_{i + 1}}(0, j)|\}
	= K_{d}\delta
	\end{aligned}
\end{equation}

Denote the new vertex generated at this iteration as $v_{new}^{i}$ which is possible to be pruned in the forthcoming iteration. Note that $\phi_{new}(T_{new}, J_{new})\in \phi^*(T_{i}, J_{i}) + \delta_{c} \mathbb{B}$. Therefore, Lemma \ref{lemma:goodvertexexists} guarantees that the vertex returned by the $best\_near\_selection$ function, denoted $v_{new}^{i'}$ is such that
$$
\CostofVertex{v_{new}^{i'}} \leq \CostofVertex{v_{new}^{i}}
$$
which implies that the pruning process will not affect (\ref{equation:flowopt}) and (\ref{equation:jumpopt}).

Define $K_{\max} := \max \{K_{c}, K_{d}\}$. Define $\Delta c_{\min}:= \min \{\Delta c_{f}, \Delta c_{g}\}$. Because of (\ref{equation:flowopt}), (\ref{equation:jumpopt}) and (\ref{equation:nfng}), then difference between the cost of $\psi^{*}= (\phi^*, u^*)$ and the cost of $\psi = (\phi, u)$ constructed by concatenating $N$ number of $\psi_{new}$  can be characterized as follows
\begin{equation}
\begin{aligned}
c(\phi) &\leq c(\phi^{*}) + N_{f}K_{c}\delta + N_{g}K_{d}\delta \leq c(\phi^{*}) + NK_{\max}\delta\\
& \leq c(\phi^{*}) + \frac{c(\phi^{*})}{\Delta c_{\min}}K_{\max}\delta = (1 + \frac{K_{\max}\cdot \delta}{\Delta c_{\min}})c(\phi^{*}).
\end{aligned}
\end{equation}
\end{proof}}

%% file: sections/numerics/main.tex
\begin{figure}[htbp]
	\centering
	\includegraphics[width= 0.7\columnwidth]{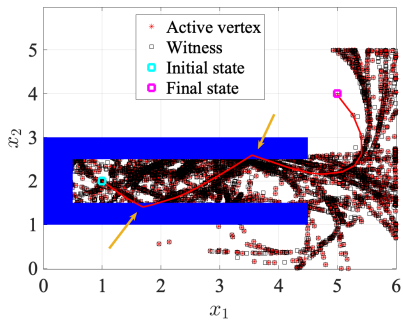}
	\caption{The motion plan generated by HySST for collision-resilient tensegrity multicopter. The blue rectangles denote the obstacles that may cause the collision with the multicopter. The yellow arrows point to the location where collision occurs.\label{fig:drone}}
\end{figure}

Algorithm \ref{algo:hybridRRT} leads to a software tool\footnote{Code at \hyperlink{https://github.com/HybridSystemsLab/hybridRRT}{https://github.com/HybridSystemsLab/hybridSST}.} to solve the optimal motion planning problems for hybrid systems. This software only requires the inputs listed in Algorithm \ref{algo:hybridRRT}. Next, HySST algorithm and this tool are illustrated in Examples \ref{example:bouncingball} and \ref{example:multicopter}.
\begin{example} (Actuated bouncing ball system)
	The simulation result in Figure \ref{fig:examplebb} shows that HySST is able to find a motion plan for the instance of optimal motion planning problem for the actuated bouncing ball system.
	\begin{figure}[htbp]
		\centering
		\subfigure[HySST\label{fig:examplehySST}]{\includegraphics[width=0.48\columnwidth]{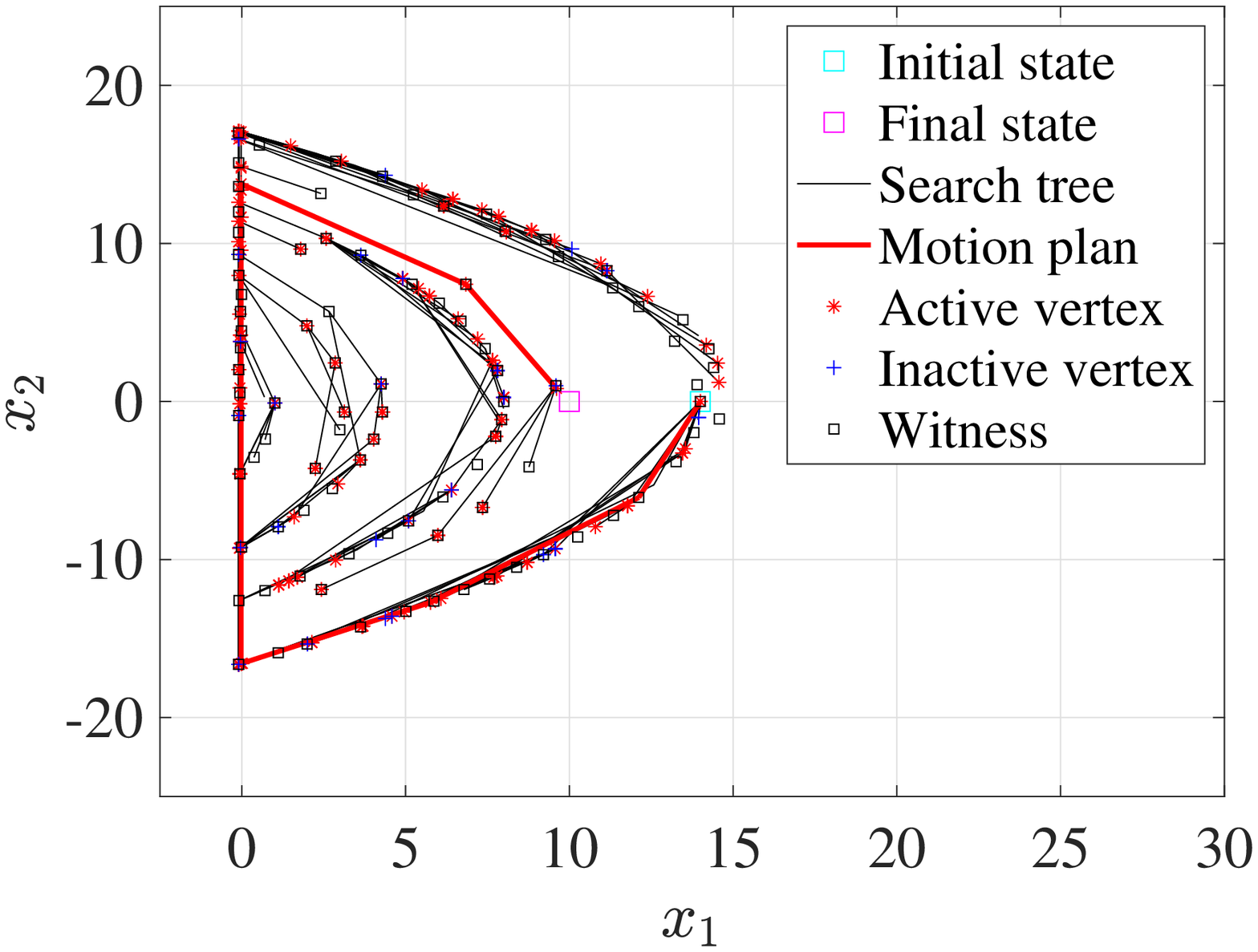}}
			\subfigure[HyRRT\label{fig:examplehyRRT}]{\includegraphics[width=0.48\columnwidth]{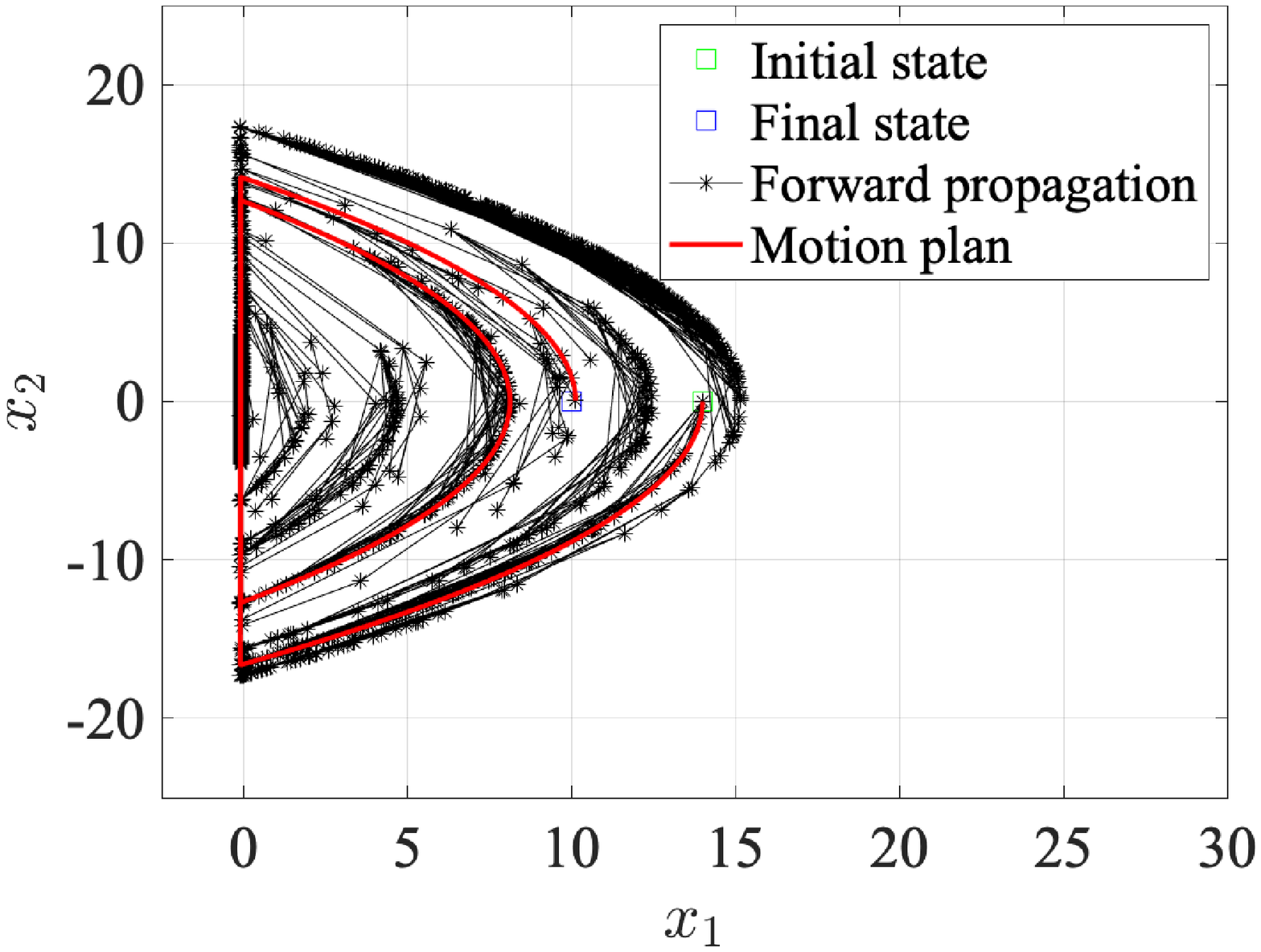}}
			\caption{Motion plans for actuated bouncing ball example solved by HySST and HyRRT in \cite{wang2022rapidly}\label{fig:examplebb}.}
	\end{figure}
The simulation is implemented in MATLAB and processed by a $3.5$ GHz Intel Core i7 processor. Both HySST and HyRRT are run for $20$ times to solve the same problem. The HySST creates  $154$ active vertices and $35$ inactive vertices and takes $3.30$ seconds, while HyRRT creates $660$ vertices in total and takes $18.4$ seconds on average. As is shown in Figure \ref{fig:examplehySST}, only one jump occurs in the motion plans generated by HySST.  Compared to the motion plans generated by HyRRT in Figure \ref{fig:examplehyRRT} where multiple jumps occur, the motion plan generated by HySST takes less hybrid time.
\end{example}
\begin{example} (Collision-resilient tensegrity multicopter)
	\ifbool{conf}{}{{\color{red} In this example, the restitution coefficient in \ref{eq:vn} is set as $0.43$ and the constant $\kappa$ in (\ref{eq:vt}) is set as $0.20$.} }The simulation result in Figure \ref{fig:drone} shows that HySST is able to ultilize the collision with the wall to decrease the hybrid time of the motion plan for multicopter. This simulation is computed on the same computation platform as the previous example, and the simulation for this problem takes $54.7$ seconds and creates $2094$ active vertices on average.
\end{example}

%% file: sections/conclusion/main.tex
In this paper, a HySST algorithm is proposed to solve optimal motion planning problems for hybrid systems. The proposed algorithm is illustrated in the bouncing ball and multicopter examples and the results show its capacity to solve the problem. In addition, this paper provides a result showing HySST algorithm is asymptotically near optimal under mild assumptions. 